\documentclass[letterpaper]{article} 
\usepackage[preprint]{aaai2027}  
\usepackage[hyphens]{url}  
\usepackage{graphicx} 
\usepackage{natbib}  
\usepackage{caption} 
\usepackage{booktabs}
\usepackage{multirow}
\usepackage{amsmath}
\usepackage{amssymb}
\usepackage{xspace}
\usepackage{tcolorbox}
\tcbuselibrary{breakable,skins}
\usepackage[capitalize]{cleveref} 
\crefname{section}{Sec.}{Secs.}
\Crefname{section}{Section}{Sections}
\crefname{table}{Tab.}{Tabs.}
\Crefname{table}{Table}{Tables}
\crefname{figure}{Fig.}{Figs.}
\Crefname{figure}{Figure}{Figures}

\makeatletter
\DeclareRobustCommand\onedot{\futurelet\@let@token\@onedot}
\def\@onedot{\ifx\@let@token.\else.\null\fi\xspace}

\makeatother

\definecolor{modcolor}{RGB}{160, 70, 0}
\definecolor{dacolor}{RGB}{148,103,189}
\definecolor{dabgcolor}{RGB}{248,242,255}
\definecolor{tracolor}{RGB}{58,105,185}
\definecolor{trabgcolor}{RGB}{236,243,255}
\definecolor{gacolor}{RGB}{75,25,115}
\definecolor{gabgcolor}{RGB}{243,232,255}
\newtcolorbox{daprompt}[1]{enhanced,title={\textbf{#1}},fonttitle=\bfseries\small,
  colback=dabgcolor,colframe=dacolor!80,colbacktitle=dacolor,coltitle=white,
  sharp corners,boxrule=0.6pt,left=6pt,right=6pt,top=4pt,bottom=4pt,breakable}
\newtcolorbox{traprompt}[1]{enhanced,title={\textbf{#1}},fonttitle=\bfseries\small,
  colback=trabgcolor,colframe=tracolor!80,colbacktitle=tracolor,coltitle=white,
  sharp corners,boxrule=0.6pt,left=6pt,right=6pt,top=4pt,bottom=4pt,breakable}
\newtcolorbox{gaprompt}[1]{enhanced,title={\textbf{#1}},fonttitle=\bfseries\small,
  colback=gabgcolor,colframe=gacolor!80,colbacktitle=gacolor,coltitle=white,
  sharp corners,boxrule=0.6pt,left=6pt,right=6pt,top=4pt,bottom=4pt,breakable}

\newlength{\fullcolfig}
\title{OPERA: A Unified Omnimodal Progressive Spatio-Temporal Reasoning Agent for Referring Video Segmentation}
\author{
    Jingchen Ni\equalcontrib\textsuperscript{\rm 1},
    Yuji Wang\equalcontrib\textsuperscript{\rm 1},
    Shannan Yan\equalcontrib\textsuperscript{\rm 1},
    Haoru Li\textsuperscript{\rm 2},
    Sitong Chen\textsuperscript{\rm 3},
    Chun Yuan\corresponding\textsuperscript{\rm 1}
}
\affiliations{
    \textsuperscript{\rm 1}Tsinghua University \quad
    \textsuperscript{\rm 2}University of California San Diego \quad
    \textsuperscript{\rm 3}ETH Zurich\\
    \{njc24, yuji-wan24, ysn24\}@mails.tsinghua.edu.cn,
    yuanc@sz.tsinghua.edu.cn
}

\begin{document}

\maketitle

\begin{abstract}
  Referring video segmentation with heterogeneous multimodal queries---spanning text, audio, and reference images---demands both robust cross-modal understanding and precise spatio-temporal reasoning. We propose \textbf{OPERA} (\textbf{O}mnimodal \textbf{P}rogressive spatio-t\textbf{E}mporal \textbf{R}easoning \textbf{A}gent), a unified reasoning agent built on a single MLLM that performs \emph{dual-axis progressive reasoning} via three specialized stages. Along the \emph{temporal axis}, a \emph{Temporal Reasoning Agent} narrows the frame search space through coarse-to-fine filtering to identify the most informative key frame. Along the \emph{spatial axis}, a \emph{Distillation Agent} establishes \emph{what} to locate via cross-modal semantic distillation, and a \emph{Grounding Agent} enhanced with GRPO determines \emph{where} the target appears, with dense mask propagation completing the pixel-level output. OPERA sets a new state of the art on OmniAVS and Ref-AVS and transfers zero-shot to standard referring video segmentation benchmarks.
\end{abstract}

\section{Introduction}
\label[section]{sec:intro}

Human perception of the physical world is inherently omnimodal, seamlessly fusing visual, textual, and audio or speech cues \cite{imagebind}. Inspired by this, omnimodal segmentation has emerged, mapping fused multimodal semantics to pixel-level segmentation and understanding of target objects in complex scenes \cite{omniavs,qwen3omni}. Unlike single-modal \cite{segformer,deeplab,wang2024convolution,ni2026fclcod} or simple cross-modal tasks \cite{iterprime,lavt,ni2025semantic}, it breaks the information barrier between heterogeneous modalities, with broad applications in autonomous driving \cite{uniad}, medical image analysis \cite{medicalsam2}, and human-computer interaction \cite{kirillov2023sam,groundedsam,wang2026deskcraft}.

\begin{figure}[!t]
  \centering
  \includegraphics[width=\linewidth]{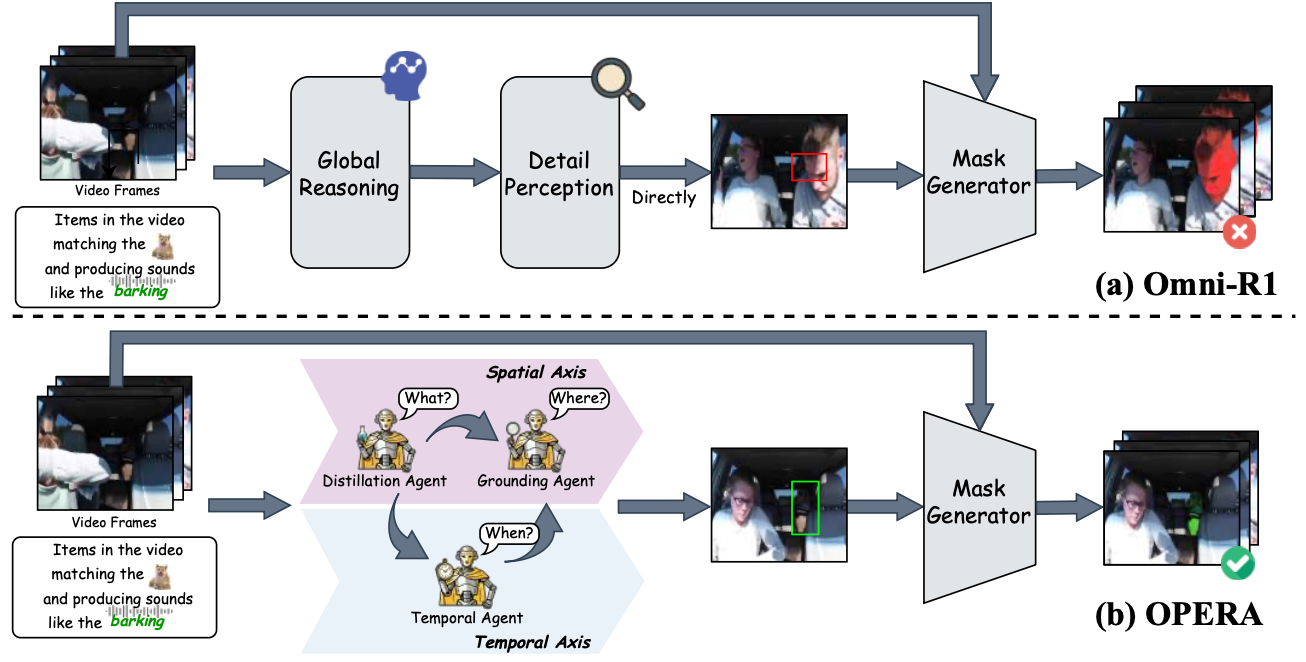}
  \caption{Comparison of Omni-R1 (a) and OPERA (b) on an omnimodal query. Omni-R1 selects a keyframe and localizes the target in two sequential LLM calls without progressive filtering, so an occluded or motion-blurred frame yields mis-localization that propagates into erroneous masks. OPERA instead reasons progressively along two axes: the \emph{temporal axis} narrows to the most informative keyframe by coarse-to-fine filtering (Temporal Agent), while the \emph{spatial axis} fixes \emph{what} to locate (Distillation Agent) and \emph{where} it appears (Grounding Agent).}
  \label[figure]{fig:teaser}
\end{figure}

Existing approaches to this problem fall short along two complementary reasoning axes (Fig.~\ref{fig:teaser}) \cite{zhongomni}. Along the \emph{temporal axis}, current methods \cite{sam2love} select grounding frames naively---relying on fixed heuristics such as the first frame or uniform sampling---without any coarse-to-fine reasoning over temporal structure. This leads to grounding on frames where the target is occluded or ambiguous, causing localization errors that propagate throughout the video. Along the \emph{spatial axis}, existing models \cite{segzero,wang2025vg} tend to jump directly from raw multimodal inputs to segmentation outputs, lacking a structured reasoning chain that first establishes \emph{what} to locate (semantic understanding), then determines \emph{where} it appears (region-level grounding), and finally produces pixel-level masks.

We propose \textbf{OPERA} (\textbf{O}mnimodal \textbf{P}rogressive spatio-t\textbf{E}mporal \textbf{R}easoning \textbf{A}gent), a unified reasoning agent for referring video segmentation with modular task decomposition \cite{fang2025cognitive,pan2026natural,yan2026adamem}. Three specialized stages share one MLLM backbone. The \emph{Temporal Reasoning Agent} progressively filters candidate frames from coarse to fine to select the most informative key frame for grounding. The \emph{Distillation Agent} converts non-textual cues into unified textual representations for downstream reasoning. The \emph{Grounding Agent} uses GRPO-enhanced chain-of-thought reasoning to localize the target, followed by dense mask propagation. Our core contributions include:

\begin{itemize}
    \item \textbf{A Unified Progressive Reasoning Agent.} OPERA decomposes referring video segmentation into three functionally specialized stages sharing one MLLM backbone, each resolving a single sub-problem and passing structured outputs to the next, so that all eight combinations of text, speech, sound, and image references are handled without modality-specific heads.

    \item \textbf{Dual-Axis Progressive Reasoning.} The temporal axis narrows the candidate frame set from uniform sampling through vision-language filtering to LLM-based reranking; the spatial axis separates \emph{what} to locate from \emph{where} it appears, yielding a coarse-to-fine reasoning chain along both axes.

    \item \textbf{State-of-the-Art Performance.} OPERA sets a new state of the art on OmniAVS and Ref-AVS and transfers zero-shot to standard referring video segmentation benchmarks, with controlled ablations attributing the gain to the reasoning protocol rather than to the backbone.
\end{itemize}

\section{Related Work}

\textbf{Omnimodal Referring Segmentation.} Omnimodal referring segmentation extends multimodal segmentation by admitting reference signals from several modalities at once. For text, the RefCOCO series (RefCOCO/+/g) \cite{refcoco} targets static images, while Refer-YouTube-VOS \cite{ReferYoutube} and MeViS \cite{mevis} carry text-guided referring segmentation into the temporal domain, with methods such as ReferFormer \cite{referformer,botach2022end,hui2023language,lan2024bifit,luo2023soc,visa,bai2024one} enabling cross-frame reasoning. Related localization tasks include cross-view correspondence~\cite{yan2026crossview} and temporal forgery localization~\cite{ni2026mgrwkv}. For audio, Ref-AVS \cite{refavs} couples audio cues with textual instructions for joint audio--text guidance, and SAM2-LOVE \cite{sam2love} improves segmentation through multimodal fusion and adaptive optimization. OmniAVS \cite{omniavs} is the representative omnimodal benchmark, supporting eight flexible expression types over text, speech, sound, and image through the OISA framework.

\textbf{Omni-Modal Foundation Models.} ImageBind \cite{imagebind} and LanguageBind \cite{languagebind} learn shared, typically image-centered embeddings for discriminative alignment rather than generation or complex reasoning. Recent omni-modal models \cite{qwen3omni,gemini,ming-omni} unify text, image, and audio for end-to-end cross-modal fusion, spatial--temporal synchronization, and real-time interaction. Reinforcement learning (RL) \cite{zhongomni,r1omni,zheng2024villa,wang2026rolling,wang2026measure} further enhances multimodal reasoning, exemplified by Omni-R1's RL-based keyframe selection and instruction rewriting for efficient single-epoch Ref-AVS training. OPERA and Omni-R1 rewrite inputs for distinct purposes: Omni-R1 bridges low-resolution reasoning and high-resolution grounding along the \emph{resolution and compute} axis; OPERA translates non-textual cues into groundable source and appearance constraints along the \emph{evidence} axis of \emph{what}, \emph{when}, and \emph{where}. These mechanisms are complementary, and our Distillation Agent is not a counterpart of Omni-R1's rewriter. Efficiency advances include quantization-aware training~\cite{lv2026reasoningqat} and speculative decoding for diffusion language models~\cite{cui2026simsd}.

\begin{figure*}[!t]
  \centering
  \includegraphics[width=0.93\linewidth]{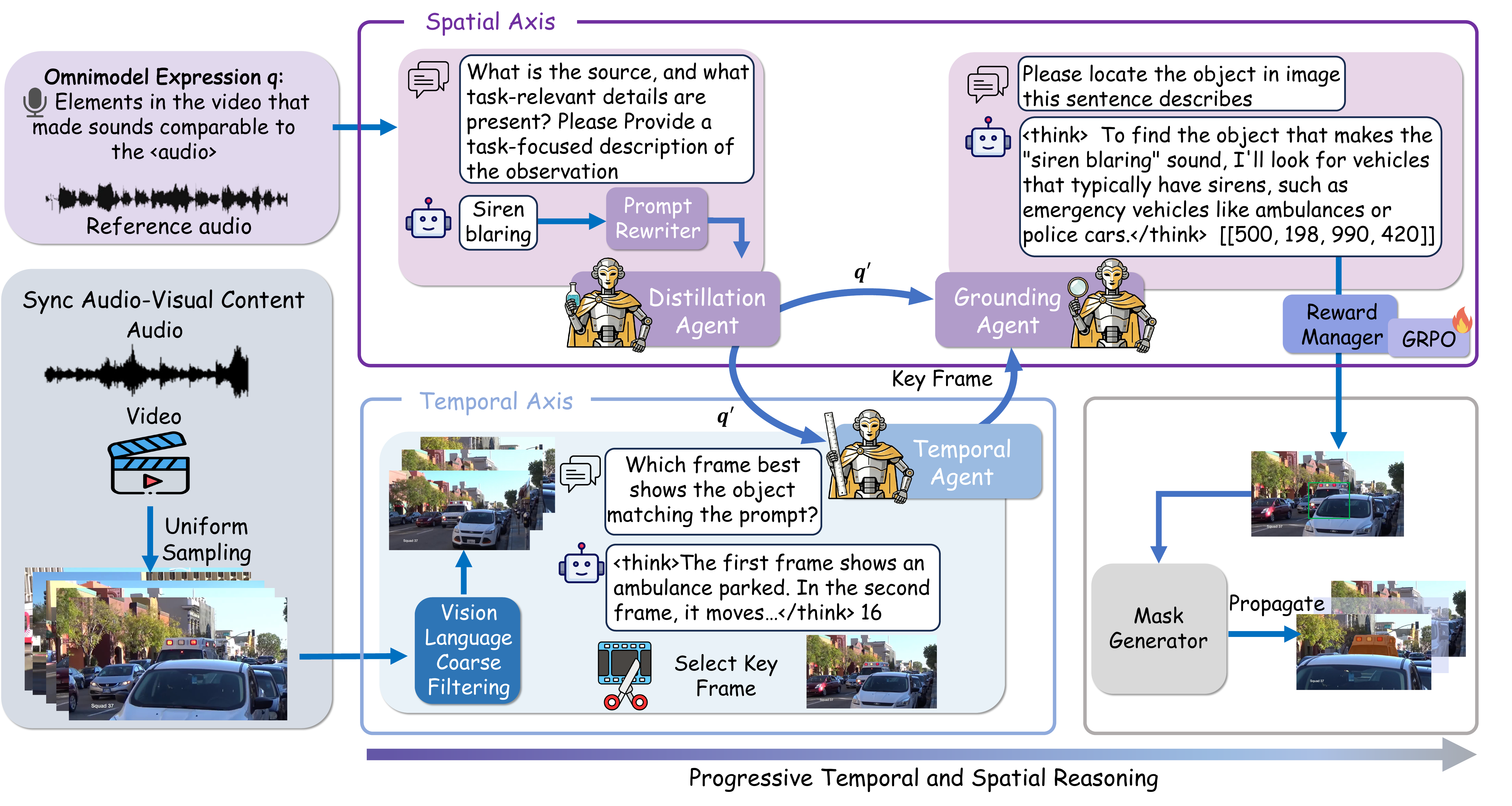}
  \caption{Overview of OPERA using a text+sound+image query as an example. Along the \textit{spatial axis}, the Distillation Agent (\S\,3.2) converts non-textual cues into concise textual hints, and the Grounding Agent (\S\,3.4) localizes the target on the selected key frame via chain-of-thought reasoning optimized with GRPO. Along the \textit{temporal axis}, the Temporal Reasoning Agent (\S\,3.3) progressively narrows the candidate frame set through uniform sampling, vision-language coarse filtering, and LLM-based semantic reranking to select a single key frame. Predicted bounding boxes are then passed to the Mask Generator for dense pixel-level propagation across all frames.}
  \label[figure]{fig:framework}
\end{figure*}

\section{Method}

\subsection{Problem Formulation and Overview}

We consider multimodal referring video segmentation: given a video \(V = \{I_t\}_{t=1}^{T}\) and a heterogeneous query, the goal is to predict a per-frame binary mask sequence \(M = \{m_t \in \{0,1\}^{H \times W}\}_{t=1}^{T}\) delineating the target. Each query pairs a mandatory language cue \(\ell \in \{\texttt{text},\,\texttt{speech}\}\) with optional supplementary modalities \(s \subseteq \{\texttt{sound},\,\texttt{image}\}\), so the query type space \(\mathcal{T} = \{(\ell,\,s)\}\) contains \(|\mathcal{T}| = 8\) combinations. Depending on \(\tau\), the query \(q_\tau\) may include a textual expression \(x\), a speech signal \(a_s\), an environmental sound clip \(a_e\), and/or a reference image \(r^{\mathrm{img}}\). The task is to learn a mapping from \((V, q_\tau)\) to \(M\) that generalizes across all \(\tau \in \mathcal{T}\) without modality-specific architectures.

We address this with OPERA (Fig.~\ref{fig:framework}). Three functionally specialized stages are instantiated from a single MLLM backbone: a \emph{Distillation Agent} (\S\,3.2) establishes \emph{what} to locate via cross-modal semantic distillation; a \emph{Temporal Reasoning Agent} (\S\,3.3) selects \emph{when} to ground via three-level progressive frame selection; and a \emph{Grounding Agent} (\S\,3.4) determines \emph{where} the target appears via GRPO-enhanced chain-of-thought reasoning, after which a dense propagation module (\S\,3.5) turns the predicted boxes into per-frame masks. The first and third operate along the spatial axis and the second along the temporal axis, realizing dual-axis coarse-to-fine reasoning as one deterministic chain. Fig.~\ref{fig:framework} traces a query with \(\tau = (\texttt{text},\,\{\texttt{sound},\,\texttt{image}\})\), \emph{i.e.}, \(q_\tau = (x,\, a_e,\, r^{\mathrm{img}})\), through every stage.

\subsection{Distillation Agent}

Along the \emph{spatial axis}, the \emph{Distillation Agent} handles the first step of progressive reasoning: establishing \emph{what} to locate. Non-textual query cues---such as environmental sounds or reference images---lack explicit semantic alignment with grounding instructions, rendering direct multimodal grounding error-prone. To address this, we introduce a two-stage distillation mechanism that converts auxiliary modalities into concise textual hints before constructing a unified grounding prompt.

\textbf{Cross-Modal Summarization.} In the first stage, the MLLM (Qwen2.5-Omni) converts each non-textual auxiliary modality into a concise textual representation. Speech \(a_s\) is transcribed into a text hint \(h^{\mathrm{sp}}\) via native speech understanding; environmental sound \(a_e\) yields a compact description \(h^{\mathrm{snd}}\) of the likely source under a minimal-hint prompt, \emph{e.g.}, ``dog barking''; and a reference image \(r^{\mathrm{img}}\) yields a caption \(h^{\mathrm{img}}\) conditioned on the query text \(x\), \emph{e.g.}, ``acoustic guitar with orange body''. Each hint retains the discriminative information needed for target identification while discarding modality-specific redundancy.

\textbf{Prompt Rewriting.} In the second stage, the hints are injected into a unified grounding prompt \(q'\) under a \emph{minimal-information principle}: for each modality set \(\tau\), the rewriter applies the least invasive transformation that preserves core evidence while suppressing modality interference. For text-with-sound queries it strips literal audio placeholders (\emph{e.g.}, ``\texttt{<audio>}'') together with the prepositional phrases referencing them, then appends a constraint directing the model to the physical object capable of producing \(h^{\mathrm{snd}}\). For text-with-image queries the image placeholder is replaced by the caption \(h^{\mathrm{img}}\). When the audio cue is speech, the raw signal \(a_s\) is retained as the primary input and processed directly, with hints from image or non-speech audio appended as verification constraints.

\subsection{Temporal Reasoning Agent}

Along the \emph{temporal axis}, the \emph{Temporal Reasoning Agent} decides \emph{when} to ground---selecting the frame that provides the most reliable basis for downstream spatial localization. Grounding on an occluded or ambiguous frame propagates localization errors across the entire video, so precise frame selection is critical. The agent implements a three-level hierarchical strategy: the first two levels use lightweight encoders for efficient cross-modal filtering, while the third level invokes an LLM for deep semantic reranking.

\textbf{Level 1: Uniform Candidate Sampling.} Given the full video with \(T\) frames, we first construct a candidate set by uniformly sampling \(K\) frames:
\begin{equation}
\mathcal{F} = \{I_{t_k}\}_{k=1}^{K}, \quad t_k = \left\lfloor \frac{(k-1)(T-1)}{K-1} \right\rfloor.
\end{equation}

\begin{table*}[!t]
  \caption{Comparison with state-of-the-art methods on OmniAVS. $\mathcal{J}\&\mathcal{F}$ is the default metric; ``All'' reports the weighted $\mathcal{J}\&\mathcal{F}$ across all samples.}
  \label[table]{tab:omniavs}
  \centering
  {\small
    \setlength{\tabcolsep}{6pt}
    \renewcommand{\arraystretch}{1.0}
    \begin{tabular}{@{}lccccccccc@{}}
      \toprule
      Method & All & I & II & III & IV & V & VI & VII & VIII \\
      \midrule
      LMPM~\cite{mevis} & 25.8 & 31.2 & 28.7 & 20.0 & 22.7 & 21.3 & 20.9 & 30.0 & 31.4 \\
      EEMC~\cite{refavs} & 29.6 & 34.4 & 32.6 & 19.6 & 26.0 & 28.0 & 24.7 & 35.6 & 36.0 \\
      MUTR~\cite{yan2024referred} & 32.3 & 35.4 & 33.3 & 28.4 & 29.8 & 26.5 & 22.8 & 41.6 & 40.5 \\
      LISA-7B~\cite{lai2024lisa} & 33.6 & 33.3 & 31.2 & 29.2 & 32.7 & 28.6 & 27.3 & 43.4 & 43.1 \\
      LISA-13B~\cite{lai2024lisa} & 36.1 & 36.4 & 32.1 & 30.4 & 35.7 & 31.6 & 30.2 & 46.7 & 45.7 \\
      OISA~\cite{omniavs} & 41.1 & 40.1 & 38.5 & 34.9 & 38.5 & 35.9 & 35.2 & \underline{52.6} & \underline{53.0} \\
      Omni-R1-7B~\cite{zhongomni} & \underline{46.6} & \textbf{55.7} & \textbf{49.5} & 38.6 & 32.3 & 15.1 & 19.7 & 14.7 & 11.4 \\
      \midrule
      OPERA-3B (ours) & 43.7 & 48.2 & 38.3 & \underline{43.3} & \underline{46.2} & \underline{45.0} & \underline{42.7} & 52.2 & 48.4 \\
      \textbf{OPERA-7B (ours)} & \textbf{53.3} & \underline{54.1} & \underline{48.4} & \textbf{61.8} & \textbf{57.0} & \textbf{59.7} & \textbf{60.1} & \textbf{66.3} & \textbf{71.8} \\
      \bottomrule
    \end{tabular}
  }
\end{table*}

\textbf{Level 2: Vision-Language Coarse Filtering.} Each candidate frame is scored against the query using a pretrained vision-language model. Let \(\mathcal{E}_T\) and \(\mathcal{E}_I\) denote its text and image encoders. The query is encoded as a text embedding and each frame as a visual embedding; cosine similarity then ranks all \(K\) candidates:
\begin{equation}
e_q = \mathcal{E}_T(x),\;
s_k = \cos\bigl(e_q, \mathcal{E}_I(I_{t_k})\bigr), k{=}1,\ldots,K.
\end{equation}
For speech queries, \(x\) is replaced by the ASR transcript \(h^{\mathrm{sp}}\) produced by the Distillation Agent (\S\,3.2). The top-\(N_c\) candidates are retained as the shortlist \(\mathcal{F}' \subset \mathcal{F}\).

\textbf{Level 3: LLM-Based Semantic Reranking.} While the vision-language encoder provides effective coarse-grained matching, it operates on shallow feature similarity and may fail to capture the semantic nuances required for accurate frame selection (\emph{e.g.}, distinguishing between an ambulance actively responding versus parked). To incorporate deeper semantic reasoning, an LLM reranks the \(N_c\) candidates. For each candidate \(I_{t_k} \in \mathcal{F}'\) we generate a temporally-aware caption \(c_k = \mathrm{LLM}_{\mathrm{cap}}(I_{t_k},\, c_{<k})\) conditioned on the captions \(c_{<k}\) of all preceding candidates, which lets the model describe inter-frame changes rather than redundantly repeating static scene elements. The captions and the query description are then presented to the LLM, which reasons over them to select the key frame \(t^* = \mathrm{LLM}_{\mathrm{rank}}(q,\, \{(k, c_k)\}_{k \in \mathcal{F}'})\). Should the reranker return an invalid index, we fall back to the highest-scoring Level-2 candidate. This hierarchical design---from \(K\) uniformly sampled candidates, to \(N_c\) coarse-filtered finalists, to one semantically selected key frame---achieves a favorable balance between computational cost and selection accuracy. The coarse-to-fine temporal refinement ensures that the final key frame exhibits clear target visibility and strong multimodal consistency.

\subsection{Grounding Agent}

Along the \emph{spatial axis}, the \emph{Grounding Agent} completes the second step of spatial progressive reasoning---localizing \emph{where} the target appears---by taking the unified instruction \(q'\) from the Distillation Agent (\S\,3.2) and the key frame \(I_{t^*}\) from the Temporal Reasoning Agent (\S\,3.3) as input. All modality combinations share a single grounding interface \(\mathcal{G}: (I_{t^*},\, q') \mapsto \{b_i\}_{i=1}^{N}\), eliminating the need for modality-specific detection heads or task-specific architectures; here \(N\) is the number of detected objects and each \(b_i = (x_1^i, y_1^i, x_2^i, y_2^i)\) is a bounding box in absolute pixel coordinates. The model input concatenates available audio tokens (included only when speech is kept in raw form), visual tokens (the key frame \(I_{t^*}\) and optionally the reference image \(r^{\mathrm{img}}\)), and text instruction tokens into a single sequence.

The model produces a structured output interleaving chain-of-thought reasoning \cite{wei2024mc,lv2025cascaded} with localization:
\begin{equation}
o = \langle\texttt{think}\rangle\; r \;\langle\texttt{/think}\rangle\; [b_1,\; b_2,\; \ldots,\; b_N],
\end{equation}
where \(r\) is a reasoning trace that makes the grounding decision interpretable, and bounding boxes are emitted in pixel coordinates consumable by downstream modules. This format is enforced during both training and inference, and unparseable outputs score zero on the format term below.

\textbf{GRPO-Enhanced Optimization.} While the base model can perform grounding natively, its unspecialized form lacks the localization precision and output regularity required for reliable segmentation. We therefore optimize the Grounding Agent via Group Relative Policy Optimization (GRPO), which generates a group of \(G\) candidate outputs per input and optimizes the policy by contrasting their relative rewards:
\begin{equation}
\mathcal{L}_{\mathrm{GRPO}} = -\mathbb{E}_{q\sim\mathcal{D}}\big[\tfrac{1}{G}\textstyle\sum_{i=1}^{G}\min(\rho_iA_i,\,c_iA_i)\big],
\end{equation}
where \(\rho_i = \pi_\theta(o_i \mid q) / \pi_{\mathrm{ref}}(o_i \mid q)\) is the probability ratio between the current and reference policies, \(c_i = \mathrm{clip}(\rho_i, 1\pm\epsilon)\) is its clipped version, \(o_i\) is the \(i\)-th generated output, and \(\epsilon\) is the clipping parameter. The advantage is obtained by group-relative normalization within the \(G\) outputs sampled for the same input, \(A_i = (R_i - \bar{R}_g)/\sigma_g\), where \(\bar{R}_g\) and \(\sigma_g\) are the mean and standard deviation of the group rewards \(\{R_j\}_{j=1}^{G}\).

The reward \(R\) combines two complementary components:
\begin{equation}
R = \alpha\, R_{\mathrm{IoU}} + (1 - \alpha)\, R_{\mathrm{fmt}},
\label{eq:reward}
\end{equation}
where \(\alpha\) balances localization quality against format compliance. The localization reward \(R_{\mathrm{IoU}}\) is the F1 score over IoU-matched boxes: a prediction counts as a true positive when \(\mathrm{IoU} \geq \delta\) against a ground-truth box and as a false positive otherwise, with precision and recall taken over the \(N_p\) predicted and \(N_g\) ground-truth boxes; we use \(\delta = 0.5\). The format reward \(R_{\mathrm{fmt}}\) awards one half for a complete \(\langle\texttt{think}\rangle\ldots\langle\texttt{/think}\rangle\) block and one half for a parseable bounding-box list, supplying the minimum structural guarantee that downstream propagation requires. Full reward specifications appear in \cref{sec:suppl-reward-spec}.



\subsection{Dense Mask Propagation}

In the final level of spatial progressive reasoning, the predicted bounding box is refined into a dense pixel-level mask and propagated across the full video. Given the bounding box \(b\) predicted on key frame \(I_{t^*}\), a video propagation module is initialized with \(b\) as a prompt on frame \(t^*\). The module stores the object representation in a memory bank \(\mathcal{M}_{t^*}\) and propagates bidirectionally---forward from \(t^*\) to frame \(T\) and backward to frame \(1\)---emitting \(m_t = \mathcal{P}(I_t,\, \mathcal{M}_{t^*})\) for every \(t \in \{1, \ldots, T\}\), where \(\mathcal{P}\) denotes the video propagation module (details in \S\,4.1). This yields the final output \(M = \{m_t\}_{t=1}^{T}\).

\section{Experiments}
\label[section]{sec:experiments}

\begin{table*}[!t]
  \caption{Results on the Ref-AVS test set. Region similarity $\mathcal{J}$, contour accuracy $\mathcal{F}$, and their average $\mathcal{J}\&\mathcal{F}$ are reported. $\dagger$~denotes results from SAM2-predicted masks based on model grounding output. For methods without reported $\mathcal{J}\&\mathcal{F}$, we compute it as $(\mathcal{J}+\mathcal{F})/2$.}
  \label[table]{tab:refavs}
  \centering
  {\small
    \setlength{\tabcolsep}{8pt}
    \renewcommand{\arraystretch}{1.0}
    \begin{tabular}{@{}lcccccc@{}}
      \toprule
      Method & \multicolumn{3}{c}{Seen} & \multicolumn{3}{c}{Unseen} \\
      \cmidrule(lr){2-4}\cmidrule(lr){5-7}
      & $\mathcal{J}\&\mathcal{F}$ & $\mathcal{J}$ & $\mathcal{F}$ & $\mathcal{J}\&\mathcal{F}$ & $\mathcal{J}$ & $\mathcal{F}$ \\
      \midrule
      AVGSegFormer\cite{gao2024avsegformer} + \texttt{text}  & 40.2 & 33.5 & 47.0 & 43.1 & 36.1 & 50.1 \\
      GAVS\cite{wang2024prompting} + \texttt{text}          & 39.4 & 28.9 & 49.8 & 39.8 & 29.8 & 49.7 \\
      ReferFormer\cite{referformer} + \texttt{audio}  & 40.7 & 31.3 & 50.1 & 39.6 & 30.4 & 48.8 \\
      R2VOS\cite{li2023robust} + \texttt{audio}        & 33.0 & 25.0 & 41.0 & 38.9 & 27.9 & 49.8 \\
      EEMC~\cite{refavs}                & 42.8 & 34.2 & 51.3 & 57.2 & 49.5 & 64.8 \\
      SAM2-LOVE~\cite{sam2love}    & 47.7 & 43.5 & 51.9 & 69.4 & 66.5 & 72.3 \\
      OISA~\cite{omniavs}         & 55.2 & 51.7 & 58.7 & 61.7 & 58.3 & 65.1 \\
      Omni-R1-7B~\cite{zhongomni} & 47.2 & 43.0 & 51.4 & \underline{74.2} & \underline{71.3} & \textbf{77.0} \\
      \midrule
      OPERA-3B (ours)      & \underline{58.6} & \underline{55.4} & \underline{61.7} & 68.3 & 67.5 & 69.1 \\
      \textbf{OPERA-7B (ours)} & \textbf{67.8} & \textbf{64.8} & \textbf{70.8} & \textbf{74.4} & \textbf{74.3} & \underline{74.6} \\
      \bottomrule
    \end{tabular}
  }
\end{table*}

\subsection{Implementation Details}
\label[subsection]{sec:impl}

\begin{table}[!t]
  \caption{Referring and reasoning video segmentation ($\mathcal{J}\&\mathcal{F}$). Rows are grouped by training setting: the upper two blocks are trained in-domain on each target benchmark, whereas OPERA is zero-shot after OmniAVS-only training, so the comparison is not apples-to-apples. Bold marks the best zero-shot and underline the best in-domain result. $\ddagger$~VISA-7B rows inherit different configurations from prior work: R-YTVOS/R-DAVIS$_{17}$ from Chat-UniVi-13B without ReVOS instruction tuning, ReVOS from VISA(IT) LLaVA-7B.}
  \label[table]{tab:rrvs}
  \centering
  {\footnotesize
    \setlength{\tabcolsep}{1pt}
    \renewcommand{\arraystretch}{1.0}
    \begin{tabular}{@{}lccc@{}}
      \toprule
      Method & R-YTVOS & R-DAVIS$_{17}$ & ReVOS \\
      \midrule
      \multicolumn{4}{@{}l}{\emph{Specialist models, in-domain fine-tuned}} \\
      ReferFormer-R50~\cite{referformer} & 62.9 & 61.1 & 14.9 \\
      MTTR (Botach et al. 2022) & 55.3 & -- & 25.5 \\
      LMPM~\cite{mevis} & -- & -- & 26.4 \\
      SOC~\cite{luo2023soc} & \underline{67.3} & 65.8 & -- \\
      MUTR~\cite{yan2024referred} & \underline{67.5} & 66.4 & -- \\
      \midrule
      \multicolumn{4}{@{}l}{\emph{MLLM-based, in-domain tuned}} \\
      LISA-7B~\cite{lai2024lisa} & 50.2 & 58.4 & 40.9 \\
      TrackGPT-7B~\cite{zhu2023tracking} & 56.4 & 63.2 & 43.6 \\
      VideoLISA-3.8B~\cite{bai2024one} & 61.7 & 67.7 & -- \\
      VISA-7B$^{\ddagger}$~\cite{visa} & 63.0 & \underline{70.4} & 47.1 \\
      OISA~\cite{omniavs} & 62.1 & 65.2 & 47.3 \\
      Omni-R1-7B~\cite{zhongomni} & -- & -- & \underline{47.6} \\
      \midrule
      \multicolumn{4}{@{}l}{\emph{Zero-shot, no in-domain training}} \\
      OPERA-3B (ours) & 61.8 & 65.9 & 52.8 \\
      \textbf{OPERA-7B (ours)} & \textbf{66.1} & \textbf{72.2} & \textbf{57.0} \\
      \bottomrule
    \end{tabular}
  }
\end{table}

\textbf{Model Architecture.} We adopt Qwen2.5-Omni~\cite{xu2025qwen25omni} (3B and 7B) as the shared MLLM backbone for all three agents. Level~2 coarse filtering uses CLIP (ViT-B/16)~\cite{radford2021learning}, with $K{=}16$ uniformly sampled frames and $N_c{=}4$ retained. Dense mask propagation uses SAM2 (Hiera-Large)~\cite{ravi2024sam2}, consistent with Omni-R1~\cite{zhongomni}.

\textbf{Training Paradigm.}
We train the Grounding Agent on 54,304 OmniAVS training samples for one epoch with full-parameter GRPO on 8 NVIDIA H20 GPUs, using a KL coefficient $\beta = 0.04$, clipping ratio $\epsilon = 0.2$, group size $G{=}8$, reward weight $\alpha = 0.8$, and an initial learning rate of $5\times10^{-7}$ under AdamW with weight decay 0.01. Rollouts are sampled at temperature 0.5 and inference is greedy, capped at 2,048 tokens; remaining settings appear in \cref{sec:suppl-grpo-config}. All other benchmarks (\S\,4.3--4.4) are evaluated zero-shot.

\subsection{Omnimodal Referring Video Segmentation}
\label[subsection]{sec:omniavs}

OmniAVS~\cite{omniavs} is a benchmark for omnimodal referring video segmentation that covers eight query splits, each combining a primary modality (text or speech) with an optional supplementary cue (environmental sound or reference image).

As shown in \cref{tab:omniavs}, OPERA-7B achieves an overall $\mathcal{J}\&\mathcal{F}$ of 53.3, surpassing OISA by 12.2 points and Omni-R1-7B by 6.7 points across all eight modality splits. Even the smaller OPERA-3B outperforms OISA, confirming that the gains stem from the dual-axis reasoning design rather than model scale alone.

The most pronounced improvements appear on the audio-dominant splits---26.9, 23.8, and 24.9 points on Splits~III, V, and VI---where targets are described through speech or environmental sound and prior methods lack structured cross-modal bridging, directly validating the Distillation Agent.

\subsection{Audio-Visual Referring Video Segmentation}
\label[subsection]{sec:refavs}

Ref-AVS~\cite{refavs} is an audio-visual referring video segmentation benchmark that evaluates generalization through a Seen split (query modalities observed during training) and an Unseen split (novel modality combinations withheld from training).

\cref{tab:refavs} reports results on Ref-AVS. OPERA-7B reaches $\mathcal{J}\&\mathcal{F}$ of 67.8 on Seen and 74.4 on Unseen, surpassing OISA by 12.6 and 12.7 points---consistent margins that rule out overfitting to known modality distributions. Against Omni-R1-7B the Seen gain reaches 20.6 points, attributable to GRPO-specialized grounding, while Unseen performance is essentially on par, so the dual-axis design does not sacrifice cross-modal generalization.

\subsection{Referring and Reasoning Video Segmentation}
\label[subsection]{sec:rrvs}

\begin{table}[!t]
  \caption{Ablation study on each OPERA component, evaluated on OmniAVS (in-domain) and Ref-AVS (zero-shot). \emph{Temporal axis}: TRA = Temporal Reasoning Agent (\S\,3.3). \emph{Spatial axis}: DA = Distillation Agent (\S\,3.2); GA = GRPO-specialized Grounding Agent (\S\,3.4). Row~IV is full OPERA.}
  \label[table]{tab:ablation-component}
  \centering
  {\footnotesize
    \setlength{\tabcolsep}{2.2pt}
    \renewcommand{\arraystretch}{1.05}
    \begin{tabular}{@{}cccccccccc@{}}
      \toprule
      \multirow{2}{*}{\textbf{ID}} & \multicolumn{1}{c}{\textbf{Temp.}} & \multicolumn{2}{c}{\textbf{Spat.}} & \multicolumn{3}{c}{\textbf{OmniAVS}} & \multicolumn{3}{c}{\textbf{Ref-AVS}} \\
      \cmidrule(lr){2-2}\cmidrule(lr){3-4}\cmidrule(lr){5-7}\cmidrule(lr){8-10}
      & TRA & DA & GA & $\mathcal{J}\&\mathcal{F}$ & $\mathcal{J}$ & $\mathcal{F}$ & $\mathcal{J}\&\mathcal{F}$ & $\mathcal{J}$ & $\mathcal{F}$ \\
      \midrule
      I   & $\times$ & $\times$ & $\times$ & 37.9 & 35.9 & 39.9 & 46.8 & 43.9 & 49.7 \\
      II  & $\times$ & $\times$ & \checkmark & 42.1 & 39.9 & 44.2 & 52.0 & 48.1 & 55.8 \\
      III & \checkmark & $\times$ & \checkmark & 44.8 & 42.4 & 47.2 & 57.2 & 53.8 & 60.6 \\
      \textbf{IV} & \checkmark & \checkmark & \checkmark & \textbf{53.3} & \textbf{51.3} & \textbf{55.3} & \textbf{67.1} & \textbf{64.8} & \textbf{69.3} \\
      \bottomrule
    \end{tabular}
  }
\end{table}

To assess zero-shot transferability, we evaluate on three standard text-query benchmarks: Ref-YouTube-VOS (R-YTVOS)~\cite{referformer}, Ref-DAVIS$_{17}$ (R-DAVIS$_{17}$)~\cite{referformer}, and ReVOS~\cite{visa}. R-YTVOS and R-DAVIS$_{17}$ test referring segmentation from natural language descriptions, while ReVOS requires reasoning over more complex expressions to identify the target. None of these datasets are used during GRPO training.

\cref{tab:rrvs} shows that OPERA-7B transfers competitively without in-domain training. This is deliberately not an apples-to-apples comparison---every baseline above is fine-tuned on its target benchmark---so the question is how much in-domain performance zero-shot transfer recovers. On R-DAVIS$_{17}$ OPERA-7B reaches 72.2, above every in-domain baseline including VISA-7B at 70.4; on R-YTVOS it scores 66.1, within 1.4 points of the fine-tuned specialists SOC and MUTR. The advantage is largest on the reasoning-heavy ReVOS, where OPERA-7B reaches 57.0 and exceeds the strongest in-domain competitor Omni-R1-7B by 9.4 points, matching the expectation that a GRPO-specialized reasoning chain helps most when the expression demands multi-step inference rather than direct visual matching.

\subsection{Ablation Study}
\label[subsection]{sec:ablation}

All ablation experiments are conducted on OmniAVS. Full OPERA-7B serves as the reference throughout.

\textbf{Effect of Each Component.}
\cref{tab:ablation-component} incrementally enables each component on OmniAVS and, zero-shot, on Ref-AVS. From the unspecialized base model, GRPO specialization adds 4.2 and 5.2 points by equipping the Grounding Agent with chain-of-thought spatial reasoning, and the Temporal Reasoning Agent a further 2.7 and 5.2 by anchoring grounding to a semantically informative key frame. The largest single gain comes from the Distillation Agent at 8.5 and 9.9 points, since without cross-modal bridging non-textual modalities severely degrade grounding. The consistent ordering across two independent benchmarks indicates that the dual-axis design captures reasoning behavior that transfers beyond the training domain.

\textbf{Key Frame Selection: Levels and Alternatives.}

\begin{table}[!t]
  \caption{Key frame selection on OmniAVS, with the Distillation Agent, Grounding Agent, and SAM2 fixed so that only the selector varies. L1: uniform candidate sampling; L2: VL coarse filtering (CLIP); L3: LLM-based semantic reranking. Time is reasoning latency per video excluding the shared SAM2 propagation; ``--'' denotes not measured. External selectors are adapted to replace only the selection stage, operating on the same rewritten query.}
  \label[table]{tab:ablation-temporal}
  \centering
  {\small
    \setlength{\tabcolsep}{5pt}
    \renewcommand{\arraystretch}{1.0}
    \begin{tabular}{@{}lcc@{}}
      \toprule
      Selector & $\mathcal{J}\&\mathcal{F}$ & Time (s) \\
      \midrule
      First frame (no selection) & 50.3 & 3.72 \\
      L1 only (uniform sampling) & 51.1 & -- \\
      L1{+}L2 (CLIP top-1) & 52.9 & 4.13 \\
      \textbf{L1{+}L2{+}L3 (full TRA)} & \textbf{53.3} & 7.69 \\
      \midrule
      VISA training-free selector~\cite{visa} & 53.4 & 23.19 \\
      AL-Ref-SAM2 GPT pivot~\cite{huang2024alrefsam2} & 52.7 & 43.21 \\
      \midrule
      Oracle within $K{=}16$ candidates & 54.1 & -- \\
      \bottomrule
    \end{tabular}
  }
\end{table}

\begin{figure}[!t]
  \centering
  \includegraphics[width=\linewidth]{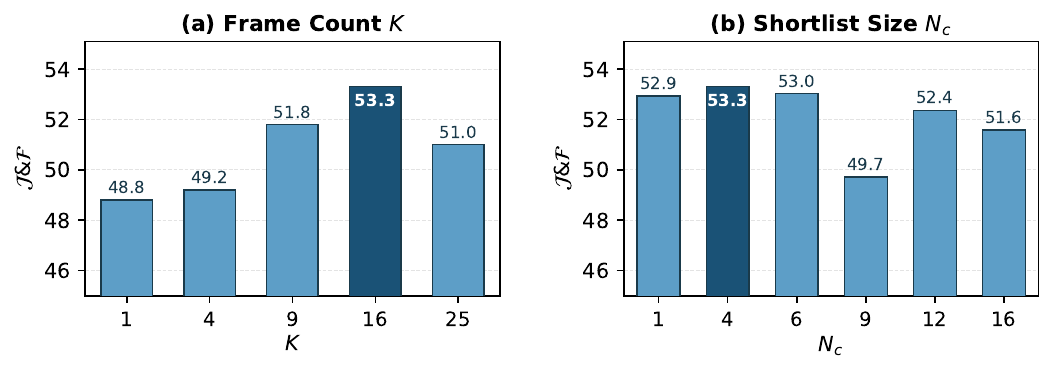}
  \caption{Sensitivity of the temporal candidate funnel on OmniAVS.
    \textbf{(a)} Uniform sample count $K$: $\mathcal{J}\&\mathcal{F}$ peaks at $K{=}16$, degrading both when candidates are too few and when excessive sampling adds near-duplicates.
    \textbf{(b)} Coarse-filter shortlist size $N_c$: $N_c{=}4$ is best, with larger shortlists making LLM reranking less reliable.}
  \label[figure]{fig:ablation-kNc}
\end{figure}

\begin{figure*}[!t]
  \centering
  \includegraphics[width=0.95\linewidth]{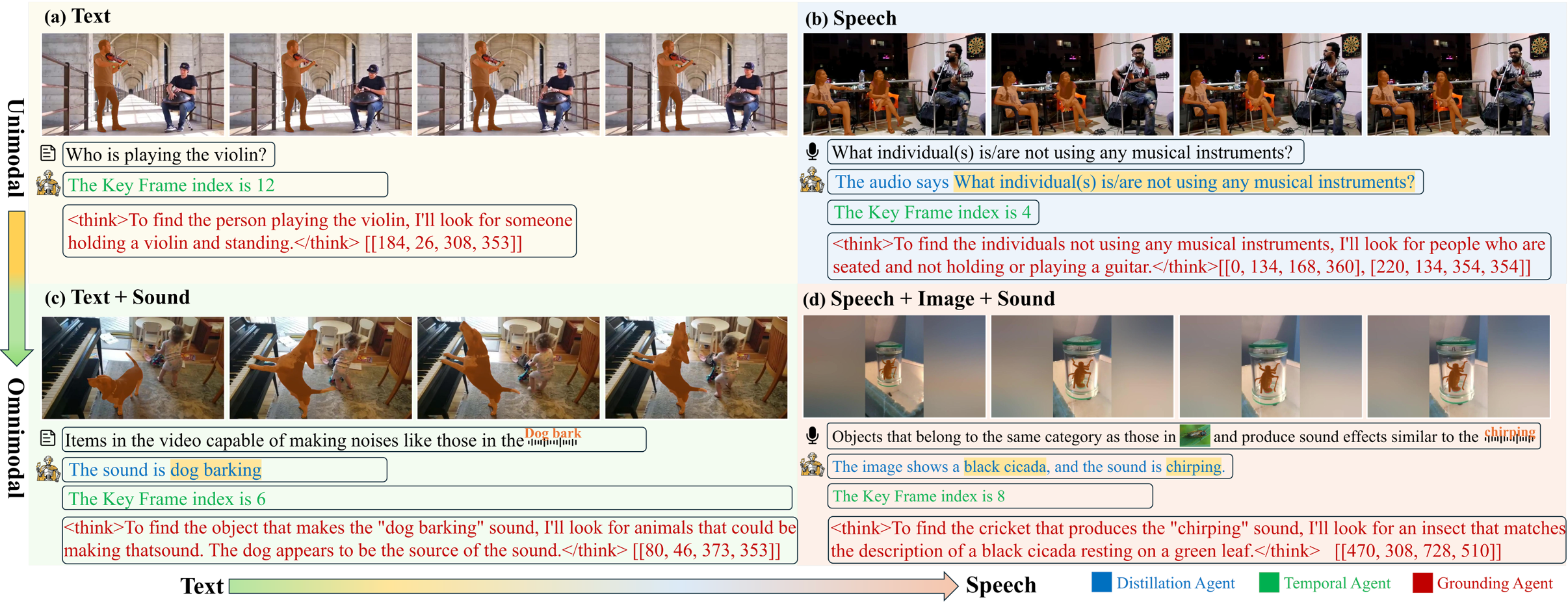}
  \caption{Qualitative results of OPERA on four examples spanning diverse query modalities. Each row shows video frames with propagated segmentation masks (orange), the selected key frame, and the chain-of-thought grounding trace with predicted bounding box.}
  \label[figure]{fig:qualitative}
\end{figure*}

\cref{tab:ablation-temporal} ablates each level of the Temporal Reasoning Agent and compares it against external selectors under an identical pipeline. Grounding on the first frame---the fixed heuristic of prior works---yields 50.3. Most of the gain comes from the two cheap levels: uniform sampling adds 0.8 points by diversifying candidates and CLIP coarse filtering a further 1.8 points. LLM semantic reranking adds 0.4 points at only 88\,ms, acting as a lightweight tie-breaker when CLIP similarity cannot separate fine-grained nuances such as an instrument being played from one merely present; we therefore do not claim that LLM reranking alone solves key frame selection.

The comparison against external selectors is more informative than these incremental gains. VISA's training-free selector reaches 53.4, indistinguishable from our 53.3, but takes 23.19\,s---about 3.0$\times$ the full TRA---while AL-Ref-SAM2's GPT-based pivot selection is slower still at 43.21\,s without improving accuracy. Crucially, an oracle picking the best of the $K{=}16$ candidates in hindsight reaches only 54.1, leaving 0.8 points of headroom for \emph{any} single-pivot policy. TRA is thus best read not as a stronger key frame predictor but as an efficient one operating near the ceiling of the single-pivot formulation itself, which places the residual error in the formulation rather than the selector.

\textbf{Temporal Hyperparameters.}
\cref{fig:ablation-kNc} sweeps the two quantities defining the candidate funnel. $\mathcal{J}\&\mathcal{F}$ peaks at $K{=}16$, falling to 51.0 at $K{=}25$ as denser sampling contributes mostly near-duplicates, and to 48.8 at $K{=}1$ when candidates are too few. The shortlist peaks at $N_c{=}4$ and degrades non-monotonically beyond it, the LLM reranker growing less reliable as it separates more semantically similar captions. We adopt $K{=}16$ and $N_c{=}4$.

\begin{table}[!t]
  \centering
  \begin{minipage}[t]{0.49\columnwidth}
    \captionof{table}{Modality-level ablation of the Distillation Agent on OmniAVS.}
    \label[table]{tab:ablation-da}
    \centering
    {\footnotesize
      \setlength{\tabcolsep}{3.8pt}
      \renewcommand{\arraystretch}{1.05}
      \begin{tabular}{@{}lccc@{}}
        \toprule
        Cond. & $\mathcal{J}\&\mathcal{F}$ & $\mathcal{J}$ & $\mathcal{F}$ \\
        \midrule
        \textbf{Full DA}  & \textbf{53.3} & \textbf{51.3} & \textbf{55.3} \\
        w/o speech        & 49.9 & 47.7 & 52.3 \\
        w/o sound         & 48.8 & 46.8 & 50.7 \\
        w/o image         & 49.8 & 47.7 & 51.9 \\
        \bottomrule
      \end{tabular}
    }
  \end{minipage}%
  \hspace{2pt}
  \begin{minipage}[t]{0.49\columnwidth}
    \captionof{table}{Ablation of GRPO reward components on OmniAVS.}
    \label[table]{tab:ablation-reward}
    \centering
    {\footnotesize
      \setlength{\tabcolsep}{3.8pt}
      \renewcommand{\arraystretch}{1.05}
      \begin{tabular}{@{}lccc@{}}
        \toprule
        Cond. & $\mathcal{J}\&\mathcal{F}$ & $\mathcal{J}$ & $\mathcal{F}$ \\
        \midrule
        w/o GRPO          & 37.9 & 35.9 & 39.9 \\
        $R_{\mathrm{IoU}}$ only & 50.3 & 48.1 & 52.4 \\
        \textbf{+$R_{\mathrm{fmt}}$} & \textbf{53.3} & \textbf{51.3} & \textbf{55.3} \\
        \bottomrule
      \end{tabular}
    }
  \end{minipage}
\end{table}

\textbf{Distillation Agent: Modality-Level Analysis.}
\cref{tab:ablation-da} disables each modality branch of the Distillation Agent at test time. Removing speech and image distillation costs 3.4 and 3.5 points on their respective query subsets; disabling sound distillation costs the most at 4.5 points, since environmental sound carries no lexical signal and bypassing distillation leaves the Grounding Agent with only raw audio embeddings. Cross-modal semantic bridging is therefore necessary for every non-textual query type.

\textbf{Grounding Agent: Reward Design Analysis.}
\cref{tab:ablation-reward} examines each reward component. The base model yields 37.9 without GRPO; the localization reward $R_{\mathrm{IoU}}$ alone lifts this to 50.3 but leaves output structure unconstrained, making reliable bounding-box extraction for SAM2 infeasible. Adding $R_{\mathrm{fmt}}$ reaches 53.3 while enforcing consistently parseable output---the minimum structural guarantee for end-to-end reliability.

\textbf{Output Format and Backbone Generality.}
Two controls separate the reasoning protocol from the backbone. First, since the reasoning text is never explicitly supervised, we do not claim that free-form chain-of-thought itself causes the gain: a \emph{bbox-only} variant that drops the reasoning trace while holding data, optimizer, reward budget, and GRPO steps fixed scores 46.7 against 53.3 for structured think{+}bbox, yet still improves on the untrained 37.9---so the benefit lies in the \emph{structured} output protocol that guarantees parseable SAM2 prompts. Second, swapping the backbone for Qwen3-Omni-30B-A3B~\cite{qwen3omni} lifts $\mathcal{J}\&\mathcal{F}$ from 43.1 to 56.2 under the same recipe, a 13.1-point gain comparable to the 15.4 on Qwen2.5-Omni-7B, indicating a backbone-agnostic harness rather than a trick specific to one model family.

\textbf{Inference Cost.}
Reasoning latency is 7.69\,s per video (DA 1.02\,s, TRA 3.97\,s, GA 2.70\,s), excluding the SAM2 propagation shared by all methods; within TRA, L3 reranking costs only 88\,ms, the rest going to CLIP filtering and captioning. This is 2.0\,s above Omni-R1-7B for 6.7 $\mathcal{J}\&\mathcal{F}$ points---an explicit accuracy--latency trade-off.

\subsection{Qualitative Results}
\label[subsection]{sec:qualitative}

Figure~\ref{fig:qualitative} traces the pipeline on four queries of increasing modality complexity. The text-only case needs no distillation, whereas the speech query demands negation reasoning over who is \emph{not} holding an instrument. The trimodal example is hardest: a reference image and a chirping cue jointly isolate a small insect that neither resolves alone. Further examples appear in \cref{sec:suppl-qualitative}.

\section{Conclusion}
We present OPERA, a unified omnimodal agent that performs dual-axis progressive reasoning for referring video segmentation: the temporal axis narrows the frame search space from coarse sampling to semantic reranking, while the spatial axis establishes what to locate and then where it appears through distillation and GRPO-optimized chain-of-thought grounding. OPERA reaches state-of-the-art results on OmniAVS and Ref-AVS and transfers zero-shot to standard benchmarks, with controlled ablations attributing the gain to the reasoning protocol rather than the backbone. Semantic distillation provides the largest gain, while efficient temporal selection approaches the single-pivot oracle ceiling and structured outputs enable reliable mask propagation.

\textbf{Limitations.} OPERA assumes the full video is available, since the Temporal Agent samples candidates from the whole clip, so streaming is out of scope. Natural extensions include a sliding-window candidate funnel, multi-pivot grounding for longer videos, and stronger omni-modal backbones evaluated under matched latency budgets.

\clearpage
\bibliography{main}
\clearpage

\raggedbottom
\appendix

\section*{Appendix}

This appendix is organised as follows.
\Cref{sec:suppl-selector} compares our key frame selector against external selectors under a controlled pipeline and reports the single-pivot oracle.
\Cref{sec:suppl-protocol} separates the contribution of the structured output protocol from that of the backbone.
\Cref{sec:suppl-impl} gives full implementation details, reward specifications, hyperparameter ablations, and prompt templates.
\Cref{sec:suppl-runtime} reports runtime and computational cost.
\Cref{sec:suppl-quant} adds quantitative results on ReVOS and MeVIS, and \cref{sec:suppl-qualitative} presents extended qualitative examples together with a failure analysis.

\section{Key Frame Selection: External Selectors and the Single-Pivot Oracle}
\label[section]{sec:suppl-selector}

\Cref{tab:ablation-temporal} of the main text compares the Temporal Reasoning Agent (TRA) against two external key frame selectors and a single-pivot oracle, with the Distillation Agent, Grounding Agent, and SAM2 held fixed on the same OmniAVS split and hardware. This section documents the protocol behind that comparison and sets out how we read the resulting numbers.

\noindent\textbf{Protocol.}
The two external selectors are adapted to replace only the selection stage: both receive the same rewritten query $q'$ produced by the Distillation Agent, and their selected frame is passed to the same Grounding Agent and the same SAM2 configuration. This isolates the choice of pivot rather than comparing full systems. All timings exclude SAM2 propagation, which is identical across rows and therefore constant.

\noindent\textbf{Reading the numbers.}
The training-free selector of VISA~\cite{visa} reaches 53.4, which is within 0.1 points of our 53.3, but requires 23.19\,s against our 7.69\,s, roughly a factor of three. The GPT-based pivot selection of AL-Ref-SAM2~\cite{huang2024alrefsam2} is slower still at 43.21\,s and does not improve accuracy. The decisive number is the oracle: selecting, in hindsight, the best of the $K{=}16$ candidates reaches only 54.1, so at most 0.8 points remain available to \emph{any} single-pivot policy. TRA should therefore be read not as a stronger key frame predictor but as an efficient one operating close to the ceiling of the single-pivot formulation itself, which places the residual error in the formulation rather than in the selector. Our temporal claim is deliberately limited in this sense: TRA is an efficient one-pivot selection policy for omnimodal inputs, and we do not claim that LLM reranking alone solves key frame selection. Promoting the top-$N$ Level-2 candidates into multiple pivots is the natural way past this ceiling and is left to future work.

\section{Output Protocol and Backbone Generality}
\label[section]{sec:suppl-protocol}

Two controls separate the contribution of the reasoning protocol from that of the backbone. Both are reported in \cref{tab:suppl-cot}.

\begin{table}[!t]
  \caption{Output protocol and backbone generality on OmniAVS. The \emph{bbox-only} row removes the reasoning trace while holding the data, optimizer, reward budget, and number of GRPO steps fixed. The lower block repeats the recipe on a different backbone.}
  \label[table]{tab:suppl-cot}
  \centering
  {\footnotesize
    \setlength{\tabcolsep}{5pt}
    \renewcommand{\arraystretch}{1.05}
    \begin{tabular}{@{}lcc@{}}
      \toprule
      Setting & GRPO & $\mathcal{J}\&\mathcal{F}$ \\
      \midrule
      Qwen2.5-Omni-7B, w/o GRPO            & --         & 37.9 \\
      Qwen2.5-Omni-7B, bbox-only / no-think & \checkmark & 46.7 \\
      \textbf{Qwen2.5-Omni-7B, full OPERA} & \checkmark & \textbf{53.3} \\
      \midrule
      Qwen3-Omni-30B-A3B~\cite{qwen3omni}  & --         & 43.1 \\
      \textbf{Qwen3-Omni-30B-A3B}          & \checkmark & \textbf{56.2} \\
      \bottomrule
    \end{tabular}
  }
\end{table}

\noindent\textbf{The gain is not attributable to free-form chain-of-thought.}
The reasoning text is never explicitly supervised, so we do not claim that free-form reasoning by itself causes the improvement. The \emph{bbox-only} variant removes the reasoning trace but keeps the data, optimizer, reward budget, and number of GRPO steps unchanged. It reaches 46.7 against 53.3 for the structured think-plus-bbox output, yet still improves substantially on the untrained 37.9. The interpretable conclusion is that the benefit lies in the \emph{structured} output protocol, which guarantees parseable prompts for SAM2, rather than in free-form reasoning as such.

\noindent\textbf{The gain is not specific to one backbone.}
Repeating the same GRPO recipe on Qwen3-Omni-30B-A3B lifts $\mathcal{J}\&\mathcal{F}$ from 43.1 to 56.2, a gain of 13.1 points that is comparable to the 15.4 points obtained on Qwen2.5-Omni-7B. This suggests that OPERA acts as a backbone-agnostic agent harness rather than a trick tied to one model family. We note that two backbones do not establish generality in full, and broader cross-backbone validation remains future work.

\section{Implementation Details}
\label[section]{sec:suppl-impl}

\subsection{GRPO Training Configuration}
\label[subsection]{sec:suppl-grpo-config}

\noindent\textbf{Hardware and optimization.}
Training uses full-parameter GRPO on 8 NVIDIA H20 GPUs with DeepSpeed ZeRO-2 offloading; no LoRA or other parameter-efficient adapters are applied.
The optimizer is AdamW with an initial learning rate of $5\times10^{-7}$, weight decay of $0.01$, and a cosine warmup ratio of $0.1$.
The effective batch size is $8$ per GPU, achieved with a per-device batch size of $1$ and $8$ gradient accumulation steps.
Training runs for $4{,}000$ optimization steps over the OmniAVS training split.

\noindent\textbf{GRPO-specific hyperparameters.}
Each input generates a group of $G=8$ candidate outputs for policy gradient estimation.
The KL penalty coefficient is $\beta=0.04$, the clipping ratio is $\epsilon=0.2$, and the reward weight is $\alpha=0.8$.
Rollout generation uses a sampling temperature of $0.5$ and a repetition penalty of $1.15$, with a maximum sequence length of $4{,}096$ tokens.
The reference policy is frozen at initialization and remains fixed throughout training.
We report the final checkpoint; no intermediate checkpoint selection is performed, and no held-out validation split is used for model selection.

\subsection{Per-Agent Inference Configuration}
\label[subsection]{sec:suppl-shared-ckpt}

\Cref{tab:suppl-ckpt} summarizes the backbone, training status, and inference decoding configuration for each OPERA module.
All MLLM-based stages---DA, TRA Level-3, and GA---share a single Qwen2.5-Omni-7B instance initialized once at inference startup; no weight copies or model reloads occur between stages.
Only GA is updated by GRPO; DA and TRA Level-3 operate on the same post-GRPO checkpoint in a frozen, zero-shot manner, inheriting the improved backbone representations without receiving any direct gradient updates for their respective tasks.

\begin{table}[!t]
  \caption{Per-agent inference configuration. All MLLM-based stages share one Qwen2.5-Omni-7B checkpoint; only GA is updated by GRPO.
  $T$ denotes the decoding temperature at inference; ``---'' indicates no autoregressive decoding.}
  \label[table]{tab:suppl-ckpt}
  \centering
  {\setlength{\tabcolsep}{3pt}
  \renewcommand{\arraystretch}{1.05}
  \resizebox{\linewidth}{!}{%
  \begin{tabular}{llllcc}
    \toprule
    \textbf{Module} & \textbf{Backbone} & \textbf{Training Status} & \textbf{Input Modalities} & \boldmath$T$ & \textbf{Max Tokens} \\
    \midrule
    Distillation Agent            & Qwen2.5-Omni-7B   & Frozen           & audio; ref.\ image; text        & 0   & 2{,}048 \\
    TRA Level-2                   & CLIP ViT-B/16     & Frozen           & frame images; text / audio      & --- & ---     \\
    TRA Level-3                   & Qwen2.5-Omni-7B   & Frozen           & candidate frames; text          & 0   & 2{,}048 \\
    Grounding Agent               & Qwen2.5-Omni-7B   & \textbf{GRPO-updated} & keyframe; rewritten query   & 0   & 2{,}048 \\
    Mask Propagation              & SAM2 Hiera-Large  & Frozen           & frame sequence; bbox prompts    & --- & ---     \\
    \bottomrule
  \end{tabular}}}
\end{table}

All MLLM-based stages adopt greedy decoding at inference, with a unified maximum output length of 2{,}048 tokens.
TRA Level-3 performs two sequential calls on the same model instance: a captioning pass that generates a one-sentence description per candidate frame, followed by a reranking pass that outputs a single keyframe index; both are capped at 2{,}048 tokens.
GRPO rollout generation during training uses $T{=}0.5$ with repetition penalty $1.15$ to encourage output diversity, as detailed in \cref{sec:suppl-grpo-config}; the greedy inference setting reported here is distinct from those training-time parameters.

\subsection{Reward Function Specification}
\label[subsection]{sec:suppl-reward-spec}

The reward $R$ combines a localization component $R_{\mathrm{IoU}}$ and a format component $R_{\mathrm{fmt}}$ as
$R = \alpha\,R_{\mathrm{IoU}} + (1{-}\alpha)\,R_{\mathrm{fmt}}$,
as defined in Eq.~\eqref{eq:reward} of the main text. This section specifies the exact implementation of each component.

\noindent\textbf{Format reward.}
The format reward $R_{\mathrm{fmt}}$ of Eq.~\eqref{eq:reward} is computed from the full model output string $s$ via two independent checks, each contributing $\tfrac{1}{2}$ to the total.
The think-chain completeness check verifies that $s$ begins with \texttt{<think>}, contains \texttt{</think>}, has non-empty content between the tags (more than $5$ characters), and includes at least $3$ characters following \texttt{</think>}.
The bounding box format check extracts a JSON-compatible substring after \texttt{</think>} via a five-step parser: (1)~locate the first \texttt{[}; (2)~advance a bracket counter to the matching \texttt{]}; (3)~evaluate the substring; (4)~verify it is a non-empty list; (5)~verify that each element is a list of exactly four numeric values.
If any step fails, the check returns $0$.
For negative samples, the check instead verifies that the output contains the string ``No target''.

The combined format reward is:
\begin{equation}
R_{\mathrm{fmt}} = \tfrac{1}{2}\,\mathbb{1}[\text{complete think chain}] + \tfrac{1}{2}\,\mathbb{1}[\text{valid bbox format}].
\end{equation}

\noindent\textbf{Handling of invalid outputs.}
Outputs that fail the bounding box check receive zero on the format term during training. At inference, a failed parse is treated as an empty prediction for that sample, and no mask is propagated. In TRA Level-3, if the reranker returns a non-integer or out-of-range frame index, we fall back to the highest-scoring Level-2 CLIP candidate, so the temporal stage always yields a valid key frame.

\subsection{Reward Ablation Studies}
\label[subsection]{sec:suppl-reward-ablation}

\begin{figure}[!t]
  \centering
  \includegraphics[width=0.82\linewidth]{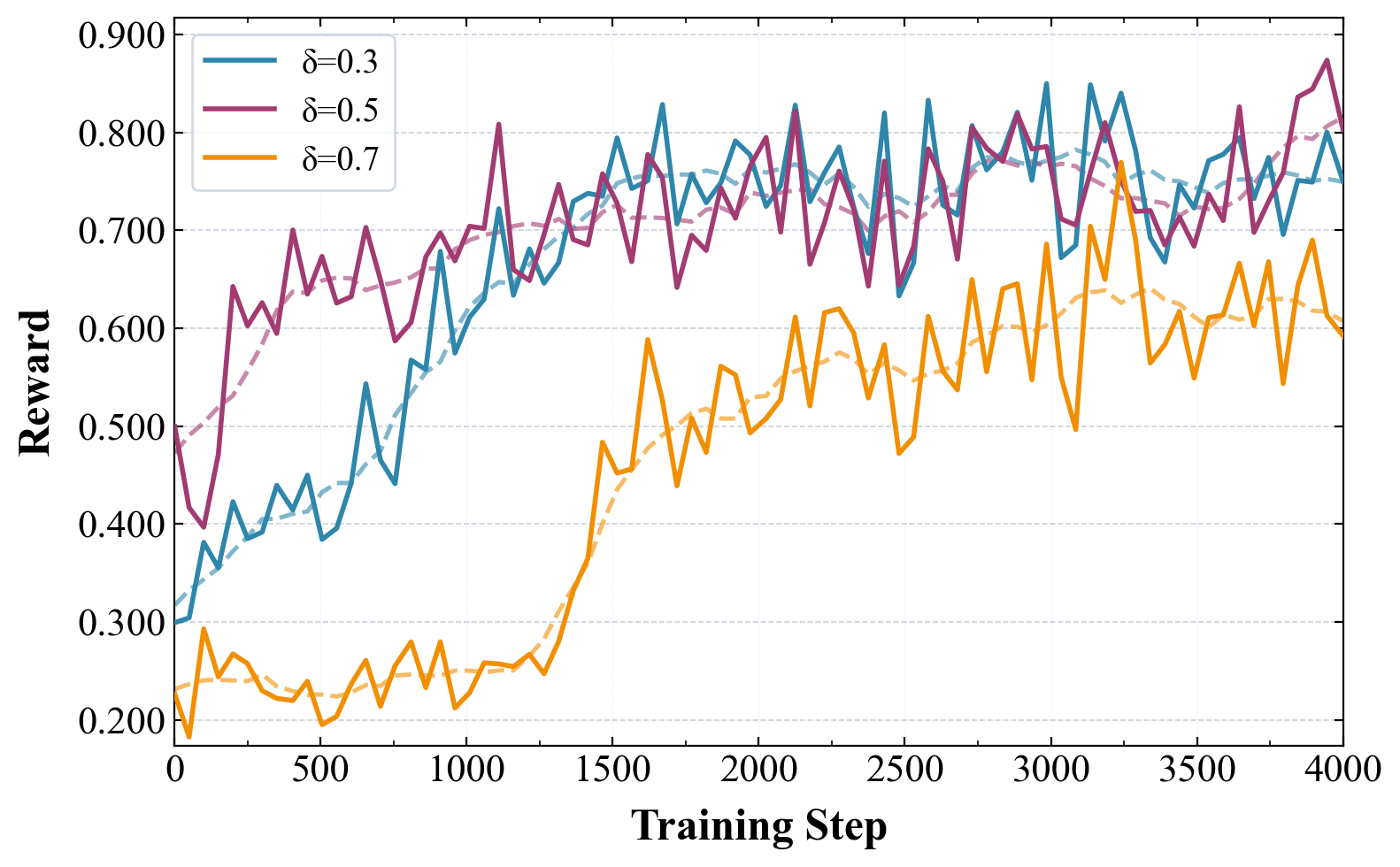}
  \caption{GRPO training reward curves for $\delta \in \{0.3, 0.5, 0.7\}$ over 4{,}000 steps (solid: raw; dashed: EMA).
  $\delta{=}0.5$ rises steadily to 0.80; $\delta{=}0.3$ reaches ${\approx}0.75$ but decouples from localization precision; $\delta{=}0.7$ stays near 0.20 for the first 1{,}500 steps due to sparse rewards.}
  \label[figure]{fig:suppl-reward-curve}
\end{figure}

\noindent\textbf{IoU threshold.}
\Cref{tab:suppl-iou-ablation} shows a 32.3-point spread in $\mathcal{J}\&\mathcal{F}$ across three $\delta$ settings, making the threshold the most sensitive hyperparameter in GRPO training.
The two failure modes are mechanistically distinct.
With $\delta{=}0.3$, the training reward converges to ${\approx}0.75$, yet $\mathcal{J}\&\mathcal{F}$ reaches only 26.7, a reward-hacking pattern in which the lenient threshold accepts over-sized boxes and decouples the objective from localization precision.
With $\delta{=}0.7$, early predictions rarely satisfy the strict criterion, collapsing the group-relative advantage and leaving the reward flat near 0.20 for the first 1{,}500 steps; the sparse signal prevents stable policy updates and yields only 21.0 $\mathcal{J}\&\mathcal{F}$.
At $\delta{=}0.5$, the threshold maintains a discriminative split throughout training: the reward rises monotonically from 0.48 to 0.80 with no plateau, achieving the best $\mathcal{J}\&\mathcal{F}$ of 53.3 (see \cref{fig:suppl-reward-curve}).

\begin{table}[!t]
  \begin{minipage}[t]{0.46\linewidth}
    \captionof{table}{IoU threshold $\delta$ ablation on OmniAVS (6{,}791 samples).
    All rows use $\alpha=0.8$ for 4{,}000 steps.}
    \label[table]{tab:suppl-iou-ablation}
    \centering
    {\footnotesize\setlength{\tabcolsep}{2.5pt}
    \renewcommand{\arraystretch}{1.0}
    \begin{tabular}{lccc}
      \toprule
      \textbf{$\delta$} & $\mathcal{J}\&\mathcal{F}$ & $\mathcal{J}$ & $\mathcal{F}$ \\
      \midrule
      0.3                      & 26.7          & 23.5          & 29.8          \\
      \textbf{0.5 (ours)}      & \textbf{53.3} & \textbf{51.3} & \textbf{55.3} \\
      0.7                      & 21.0          & 17.5          & 24.5          \\
      \bottomrule
    \end{tabular}}
  \end{minipage}
  \hfill
  \begin{minipage}[t]{0.50\linewidth}
    \captionof{table}{Reward weight $\alpha$ ablation on OmniAVS (6{,}791 samples).
    All rows use $\delta = 0.5$ for 4{,}000 steps.}
    \label[table]{tab:suppl-alpha-ablation}
    \centering
    {\footnotesize\setlength{\tabcolsep}{2.5pt}
    \renewcommand{\arraystretch}{1.0}
    \begin{tabular}{lccc}
      \toprule
      \textbf{$\alpha$} & $\mathcal{J}\&\mathcal{F}$ & $\mathcal{J}$ & $\mathcal{F}$ \\
      \midrule
      1.0                      & 50.3          & 48.1          & 52.4          \\
      \textbf{0.8 (ours)}      & \textbf{53.3} & \textbf{51.3} & \textbf{55.3} \\
      0.6                      & 37.4          & 39.2          & 35.6          \\
      0.4                      & 25.0          & 21.4          & 28.6          \\
      \bottomrule
    \end{tabular}}
  \end{minipage}
\end{table}

\noindent\textbf{Reward weight ratio.}
\Cref{tab:suppl-alpha-ablation} ablates $\alpha$, which balances the localization reward against the format reward.
Pure IoU supervision ($\alpha{=}1.0$) already yields a competitive 50.3 $\mathcal{J}\&\mathcal{F}$, confirming that the localization signal is the dominant learning driver.
Adding a modest format component at $\alpha{=}0.8$ raises performance to 53.3, a gain of 3.0 points, suggesting that enforcing parseable output syntax provides a regularizing effect that benefits downstream evaluation.
Below this optimum, performance degrades steeply: $\mathcal{J}\&\mathcal{F}$ falls to 37.4 at $\alpha{=}0.6$ and to 25.0 at $\alpha{=}0.4$, indicating that as the localization weight falls below 0.8, the weakened IoU signal allows the model to satisfy the format criterion with spatially inaccurate predictions, reproducing the reward-hacking failure observed under lenient $\delta$.

Taken together, these two ablations show that $\delta$ and $\alpha$ are the most consequential hyperparameters in the recipe, which is why both are reported explicitly rather than left to a default.

\subsection{Prompt Templates}
\label[subsection]{sec:suppl-prompts}

OPERA uses three agent-specific prompt templates, one for each functional stage. The \textbf{Distillation Agent} (DA) translates each non-textual modality into a compact textual hint via a dedicated prompt; the resulting hints are merged into the rewritten grounding query $q'$ (\cref{fig:prompt-da}).

\begin{figure}[!t]
  \centering
  \begin{minipage}{0.95\linewidth}
  \begin{daprompt}{Distillation Agent \textnormal{---} Sound}
  {\ttfamily\small
  Listen to this audio carefully.\\[3pt]
  1.~What object or creature is making this sound?\\
  2.~What are the characteristics of this sound (loud/soft, continuous/intermittent)?\\[3pt]
  Please describe concisely in one sentence, like ``a dog barking loudly'' or ``a car engine running''.
  }
  \end{daprompt}
  \vspace{4pt}
  \begin{daprompt}{Distillation Agent \textnormal{---} Speech}
  {\ttfamily\small
  Listen to this audio. A person is referring to a specific object or person in a video for visual grounding.\\
  Your task: extract ONLY the referring expression as a concise noun phrase.\\[3pt]
  Rules:\\
  \quad -- Output the noun phrase ONLY, no explanations\\
  \quad -- Keep it natural and concise, like a human would write it\\
  \quad -- Include key distinguishing attributes (color, position, action, clothing, etc.)\\[3pt]
  Examples: ``the woman in red shirt'', ``the dog barking on the left''\\[3pt]
  Referring expression:
  }
  \end{daprompt}
  \vspace{4pt}
  \begin{daprompt}{Distillation Agent \textnormal{---} Reference Image}
  {\ttfamily\small
  Given the task description: ``\{x\}''\\[3pt]
  The \textless{}image\textgreater{} in the description refers to a reference object or person.\\
  Describe what this image shows in detail, focusing on attributes relevant to the task.\\
  Include: type/category, key visual features (color, shape, gender if person).\\[3pt]
  Answer in 5--15 words:
  }
  \end{daprompt}
  \end{minipage}
  \caption{Prompt templates for the Distillation Agent. Each non-textual modality uses a dedicated prompt to produce a compact textual hint~$h$, which is subsequently merged into the rewritten grounding query~$q'$.}
  \label[figure]{fig:prompt-da}
\end{figure}

The \textbf{Temporal Reasoning Agent} Level-3 uses a two-stage cascade: a captioning prompt generates a one-sentence description per candidate frame, and a reranking prompt selects the optimal keyframe index (\cref{fig:prompt-tra}).

\begin{figure}[!t]
  \centering
  \begin{minipage}{0.95\linewidth}
  \begin{traprompt}{Temporal Reasoning Agent Level-3 \textnormal{---} Frame Captioning}
  {\ttfamily\small
  You are analyzing a video frame by frame.\\[3pt]
  Previous frame (frame \{k-1\}): \{prev\_caption\}\\[1pt]
  Current frame (frame \{k\}/\{total-1\}): [shown in image]\\[3pt]
  Describe the current frame considering what changed from the previous frame. Focus on:\\
  \quad 1.~What objects/people appear or disappear?\\
  \quad 2.~What motions or actions are happening?\\
  \quad 3.~Any changes in position or state?\\[3pt]
  Answer with ONLY 1 short sentence (max 10 words):
  }
  \end{traprompt}
  \vspace{4pt}
  \begin{traprompt}{Temporal Reasoning Agent Level-3 \textnormal{---} Frame Reranking}
  {\ttfamily\small
  You are an expert at selecting the best video frame for object localization.\\[3pt]
  Query: \{query\_desc\}\\[3pt]
  Available Video Frames (\{N\_c\} frames, in temporal order):\\
  \quad [0] \{c\_0\} (start) \quad\ldots\quad [N\_c-1] \{c\_\{N\_c-1\}\} (end)\\[3pt]
  Requirements:\\
  \quad 1.~The selected frame should clearly contain the target object.\\
  \quad 2.~The object should be visible and identifiable.\\
  \quad 3.~Consider all aspects of the query (text, audio hints, reference images).\\
  \quad 4.~Consider temporal context.\\[3pt]
  Answer with ONLY the frame index number (0--\{N\_c-1\}), nothing else:
  }
  \end{traprompt}
  \end{minipage}
  \caption{Prompt templates for TRA Level-3. The captioning prompt (top) generates a temporally-aware one-sentence description for each candidate frame; for $k{=}0$ the ``Previous frame'' line is omitted. The reranking prompt (bottom) selects the keyframe index. If the model returns a non-integer or out-of-range value, the highest-scoring Level-2 CLIP candidate is used as fallback.}
  \label[figure]{fig:prompt-tra}
\end{figure}

The \textbf{Grounding Agent} receives the rewritten query $q'$ and produces chain-of-thought reasoning before predicting bounding boxes in absolute pixel coordinates (\cref{fig:prompt-ga}).

\begin{figure}[!t]
\centering
\begin{minipage}{0.95\linewidth}
\begin{gaprompt}{Grounding Agent \textnormal{---} Localization Prompt}
{\ttfamily\small
Please locate the object this sentence describes in this image and provide bounding box(es):\\
\quad \textless{}ref\textgreater{}\{q'\}\textless{}/ref\textgreater{}\\[3pt]
Output format: Provide bounding box(es) as a list in the format [xmin, ymin, xmax, ymax],\\
where coordinates are absolute pixel values.\\
For multiple objects: [[xmin1,ymin1,xmax1,ymax1], [xmin2,ymin2,xmax2,ymax2], ...]\\[3pt]
Please think step-by-step in \textless{}think\textgreater{}\textless{}/think\textgreater{} tags first,\\
then provide your answer with bounding boxes in [] format.
}
\end{gaprompt}
\end{minipage}
\caption{Grounding Agent localization prompt. The rewritten query~$q'$ produced by the Distillation Agent is substituted at \texttt{\{q'\}} at runtime. Bounding boxes are predicted in absolute pixel coordinates and passed directly to SAM2 as spatial prompts for mask propagation.}
\label[figure]{fig:prompt-ga}
\end{figure}

\section{Runtime and Computational Cost}
\label[section]{sec:suppl-runtime}

We report per-stage latency and $\mathcal{J}\&\mathcal{F}$ for all pipeline variants on the full OmniAVS test set of 6{,}791 samples, evaluated on a single NVIDIA H20 GPU. SAM2 propagation is excluded from all timing measurements as it is constant across methods.

\noindent\textbf{GRPO delivers consistent gains at minimal overhead.}
\Cref{tab:suppl-runtime} shows that GRPO training alone lifts $\mathcal{J}\&\mathcal{F}$ from 37.9 to 42.1, a gain of 4.2 points achieved at a latency cost of only 229\,ms.
At 54.5\,ms per point, this is the lowest overhead of any pipeline component and serves as the efficiency reference for evaluating TRA and DA.

\noindent\textbf{TRA without semantic distillation is cost-inefficient.}
Adding TRA to the GRPO-trained baseline incurs 4{,}004\,ms of additional latency yet recovers only 2.7 $\mathcal{J}\&\mathcal{F}$ points overall, a cost-to-gain ratio more than twenty times worse than GRPO.
The per-category gains in \cref{tab:suppl-da-modality} explain why: for text-only queries the residual DA contribution of 0.7 points indicates that CLIP-based keyframe selection already suffices when the query is fully textual, whereas for categories with non-textual modalities the subsequent DA gains reach as high as 39.8 points. TRA without semantic distillation cannot bridge non-textual signals to visual frame retrieval.

\noindent\textbf{DA offers the best gain-to-cost ratio.}
Activating DA costs 1{,}022\,ms in direct overhead while simultaneously reducing GA latency by 675\,ms, yielding a net pipeline increase of only 443\,ms for a gain of 8.5 $\mathcal{J}\&\mathcal{F}$ points.
At 52\,ms per point, this net efficiency is comparable to that of GRPO and stands in contrast to TRA's unaided cost, confirming that DA is the primary driver of multimodal performance.

\noindent\textbf{OPERA and Omni-R1 trade latency for accuracy.}
The full pipeline runs at 7{,}693\,ms per video, 2{,}028\,ms above Omni-R1-7B, and delivers 53.3 $\mathcal{J}\&\mathcal{F}$ against 46.6.
Within TRA, Level-3 reranking accounts for only 88\,ms; the cost is dominated by CLIP filtering and candidate captioning.
Notably, the GRPO-only configuration scores 42.1 and the GRPO-plus-TRA configuration 44.8, both below Omni-R1 at 46.6; it is the addition of DA that reverses the deficit.
\begin{table}[!t]
  \caption{Component ablation with per-stage latency (ms per video) on OmniAVS.
  TRA latency is the sum of L2 (CLIP) and L3 (LLM captioning and reranking). SAM2 propagation is excluded.}
  \label[table]{tab:suppl-runtime}
  \centering
  \renewcommand{\arraystretch}{1.1}
  \resizebox{\linewidth}{!}{%
  \begin{tabular}{cccrrrrr}
    \toprule
    \textbf{TRA} & \textbf{DA} & \textbf{GA} & \textbf{DA (ms)} & \textbf{TRA (ms)} & \textbf{GA (ms)} & \textbf{Total} & $\mathcal{J}\&\mathcal{F}$ \\
    \midrule
    $\times$   & $\times$   & $\times$   & 0       & 0       & 3{,}017 & 3{,}017          & 37.9 \\
    $\times$   & $\times$   & \checkmark & 0       & 0       & 3{,}246 & 3{,}246          & 42.1 \\
    \checkmark & $\times$   & \checkmark & 0       & 3{,}877 & 3{,}373 & 7{,}250          & 44.8 \\
    \checkmark & \checkmark & \checkmark & 1{,}022 & 3{,}973 & 2{,}698 & \textbf{7{,}693} & \textbf{53.3} \\
    \midrule
    \multicolumn{3}{c}{Omni-R1-7B~\cite{zhongomni}} & \multicolumn{3}{c}{\textit{n/a}} & 5{,}665 & 46.6 \\
    \bottomrule
  \end{tabular}}
\end{table}

\begin{table}[!t]
  \caption{Distillation Agent latency and $\mathcal{J}\&\mathcal{F}$ gain per expression type.
  $\Delta\mathcal{J}\&\mathcal{F}$ compares full OPERA against the GRPO-plus-TRA configuration.
  {\color{modcolor}\texttt{sound}} and {\color{modcolor}\texttt{image}} in category names mark non-textual reference signals.}
  \label[table]{tab:suppl-da-modality}
  \centering
  {\footnotesize
  \renewcommand{\arraystretch}{1.05}
  \setlength{\tabcolsep}{6pt}
  \resizebox{\linewidth}{!}{%
  \begin{tabular}{lrr}
    \toprule
    \textbf{Category} & \textbf{DA (ms)} & \boldmath$\Delta\mathcal{J}\&\mathcal{F}$ \\
    \midrule
    \texttt{text}   & 524     & $+$0.7  \\
    \texttt{speech} & 1{,}203 & $+$6.6  \\
    \texttt{text\_{\color{modcolor}sound}}                            & 971     & $+$24.8 \\
    \texttt{speech\_{\color{modcolor}sound}}                          & 1{,}603 & $+$18.2 \\
    \texttt{text\_{\color{modcolor}image}}                            & 1{,}340 & $+$28.5 \\
    \texttt{speech\_{\color{modcolor}image}}                          & 2{,}005 & $+$39.8 \\
    \texttt{text\_{\color{modcolor}sound}\_{\color{modcolor}image}}   & 1{,}892 & $+$34.2 \\
    \texttt{speech\_{\color{modcolor}sound}\_{\color{modcolor}image}} & 2{,}429 & $+$38.9 \\
    \bottomrule
  \end{tabular}}}
\end{table}
\noindent\textbf{TRA latency is stable across query complexity.}
TRA latency rises by only 96\,ms when DA is activated, from 3{,}877 to 3{,}973\,ms, a change of 2.5\%.
This near-invariance indicates that TRA's cost is driven by video processing volume rather than by linguistic query complexity, so the overhead remains predictable as the number of input modalities grows.

\section{Additional Quantitative Results}
\label[section]{sec:suppl-quant}

\noindent\textbf{Reasoning gains dominate on ReVOS.}
\Cref{tab:suppl-revos} decomposes ReVOS performance into a Referring split, where targets are identified through direct visual descriptions, and a Reasoning split, where logical inference is required to determine the referent.
OPERA-7B achieves an overall $\mathcal{J}\&\mathcal{F}$ of 57.0, outperforming Omni-R1-7B at 47.6 by 9.4 points.
This advantage is not uniform: on the Referring split OPERA-7B scores 60.4 against 53.2, a margin of 7.2 points, whereas on the Reasoning split the gap widens to 10.7 points, 52.6 against 41.9.
The disproportionate gain on the Reasoning split is consistent with the expectation that GRPO-optimized chain-of-thought grounding helps most where expressions require multi-step inference rather than direct visual matching.
The Single-target and Multi-target splits improve by 9.0 and 8.9 points respectively, indicating that the benefit is not specific to referent cardinality.

\noindent\textbf{Gains persist at smaller scale on ReVOS.}
OPERA-3B reaches an overall 52.8, surpassing Omni-R1-7B by 5.2 points with less than half the parameter count.
The pattern holds per sub-task: on the Reasoning split OPERA-3B scores 50.4 against 41.9, and on the Referring split 54.6 against 53.2.
This cross-scale advantage points to the dual-axis reasoning design rather than to model capacity.
\begin{table}[!t]
  \caption{ReVOS per-split performance ($\mathcal{J}\&\mathcal{F}$, \%).}
  \label[table]{tab:suppl-revos}
  \centering
  {\footnotesize
  \setlength{\tabcolsep}{4pt}
  \renewcommand{\arraystretch}{1.05}
  \resizebox{\linewidth}{!}{%
  \begin{tabular}{lccccc}
    \toprule
    \textbf{Model} & \textit{Referring} & \textit{Reasoning} & \textit{Single} & \textit{Multi} & \textbf{Overall} \\
    \midrule
    Omni-R1-7B~\cite{zhongomni} & 53.2 & 41.9 & 48.3 & 46.5 & 47.6 \\
    \midrule
    OPERA-3B          & \underline{54.6} & \underline{50.4} & \underline{52.9} & \underline{52.0} & \underline{52.8} \\
    \textbf{OPERA-7B} & \textbf{60.4}    & \textbf{52.6}    & \textbf{57.3}    & \textbf{55.4}    & \textbf{57.0}    \\
    \bottomrule
  \end{tabular}}}
  \vspace{0.7em}
\end{table}

\begin{table}[!t]
  \caption{MeVIS val\_u ($\mathcal{J}\&\mathcal{F}$, \%).}
  \label[table]{tab:suppl-mevis}
  \centering
  {\footnotesize
  \setlength{\tabcolsep}{8pt}
  \renewcommand{\arraystretch}{1.05}
  \begin{tabular}{lc}
    \toprule
    \textbf{Model} & \textbf{MeVIS} \\
    \midrule
    Sa2VA-1B~\cite{hong2024sa2va}    & 53.4 \\
    Sa2VA-4B~\cite{hong2024sa2va}    & 55.4 \\
    \midrule
    Qwen2.5-Omni-7B (base)           & 33.6 \\
    Omni-R1-8B~\cite{zhongomni}      & 55.4 \\
    \midrule
    OPERA-3B          & \underline{48.8} \\
    \textbf{OPERA-7B} & \textbf{59.4} \\
    \bottomrule
  \end{tabular}}
\end{table}
\noindent\textbf{MeVIS results are scale-dependent.}
On MeVIS val\_u (\cref{tab:suppl-mevis}), OPERA-7B attains 59.4, surpassing both Sa2VA-4B and Omni-R1-8B by 4.0 points.
The base Qwen2.5-Omni-7B backbone without task-specific adaptation scores only 33.6, a gap of 25.8 points, confirming that the progressive reasoning chain rather than the backbone accounts for the result.
Unlike ReVOS, where OPERA-3B already exceeds larger competing models, on MeVIS the 3B variant scores 48.8 and falls below both Sa2VA-1B at 53.4 and Omni-R1-8B at 55.4.
This suggests that resolving motion-described temporal expressions places greater demands on model capacity than the reasoning-style expressions in ReVOS.

\section{Extended Qualitative Results and Failure Analysis}
\label[section]{sec:suppl-qualitative}

\Crefrange{fig:suppl-qual-a}{fig:suppl-qual-c} present six OPERA examples ordered by increasing modality complexity, from text-only and speech queries through dual-modal combinations to trimodal inputs, covering five of the eight OmniAVS query types.
Each example traces the full pipeline: green boxes show Distillation Agent hints; blue boxes show TRA outputs including L2 candidate frames and L3 reranking chain-of-thought; red boxes show the Grounding Agent's reasoning trace and predicted bounding boxes.
In none of the six cases does the selected key frame fall at the start of the clip: the Level-2 shortlists are spread across the video and the Level-3 reranker settles on indices such as 12, 26, and 33. This is the per-video form of the 3.0-point first-frame gap reported in \cref{tab:ablation-temporal}.

\noindent\textbf{Failure analysis.}
Because OPERA is a deterministic chain, an error at one stage is inherited by those that follow. Four failure modes recur. The Distillation Agent may discard discriminative cues when the speech is ambiguous or the reference image admits several readings. The Temporal Reasoning Agent may settle on a cluttered pivot when candidates score similarly under CLIP. The Grounding Agent may confuse instances of the same class. SAM2 may drift after a long target absence.

A representative case contains two instruments of the same class: the Distillation Agent binds the reference image to the wrong instance, and because the pipeline commits to a single key frame, SAM2 propagates that identity for the rest of the video. Multi-pivot grounding is the natural remedy.

\par\noindent
\begin{minipage}[t][\fullcolfig][t]{\linewidth}
  \centering
  \includegraphics[width=\linewidth]{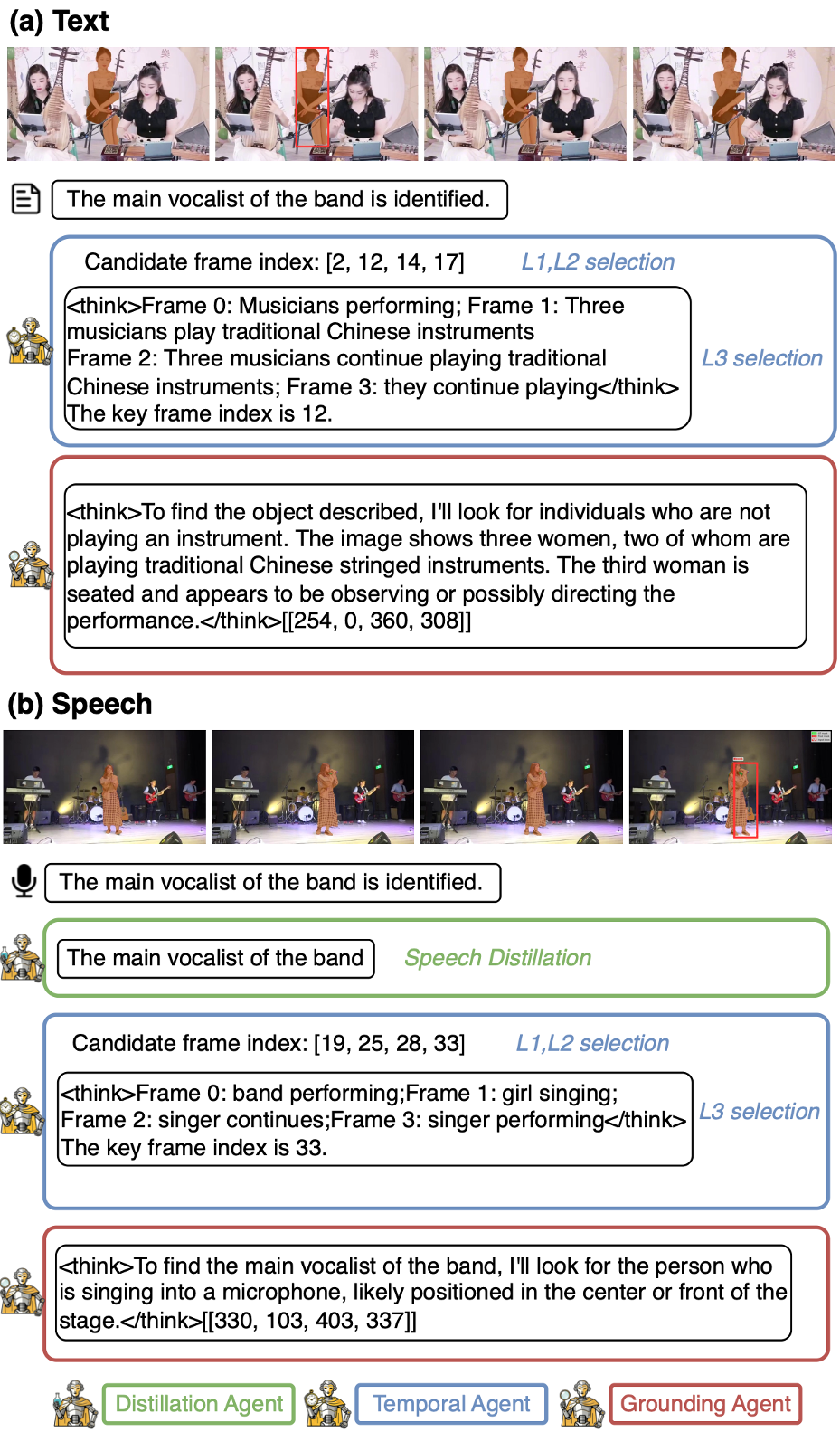}
  \captionof{figure}{Qualitative examples of OPERA on text-only (a) and speech (b) queries.
  In (a), no DA hint is needed; the TRA directly selects the key frame and the GA localizes the target.
  In (b), the DA extracts the referring expression from speech via speech distillation before the TRA and GA stages.}
  \label[figure]{fig:suppl-qual-a}
\end{minipage}\par

\par\noindent
\begin{minipage}[t][\fullcolfig][t]{\linewidth}
  \centering
  \includegraphics[width=\linewidth]{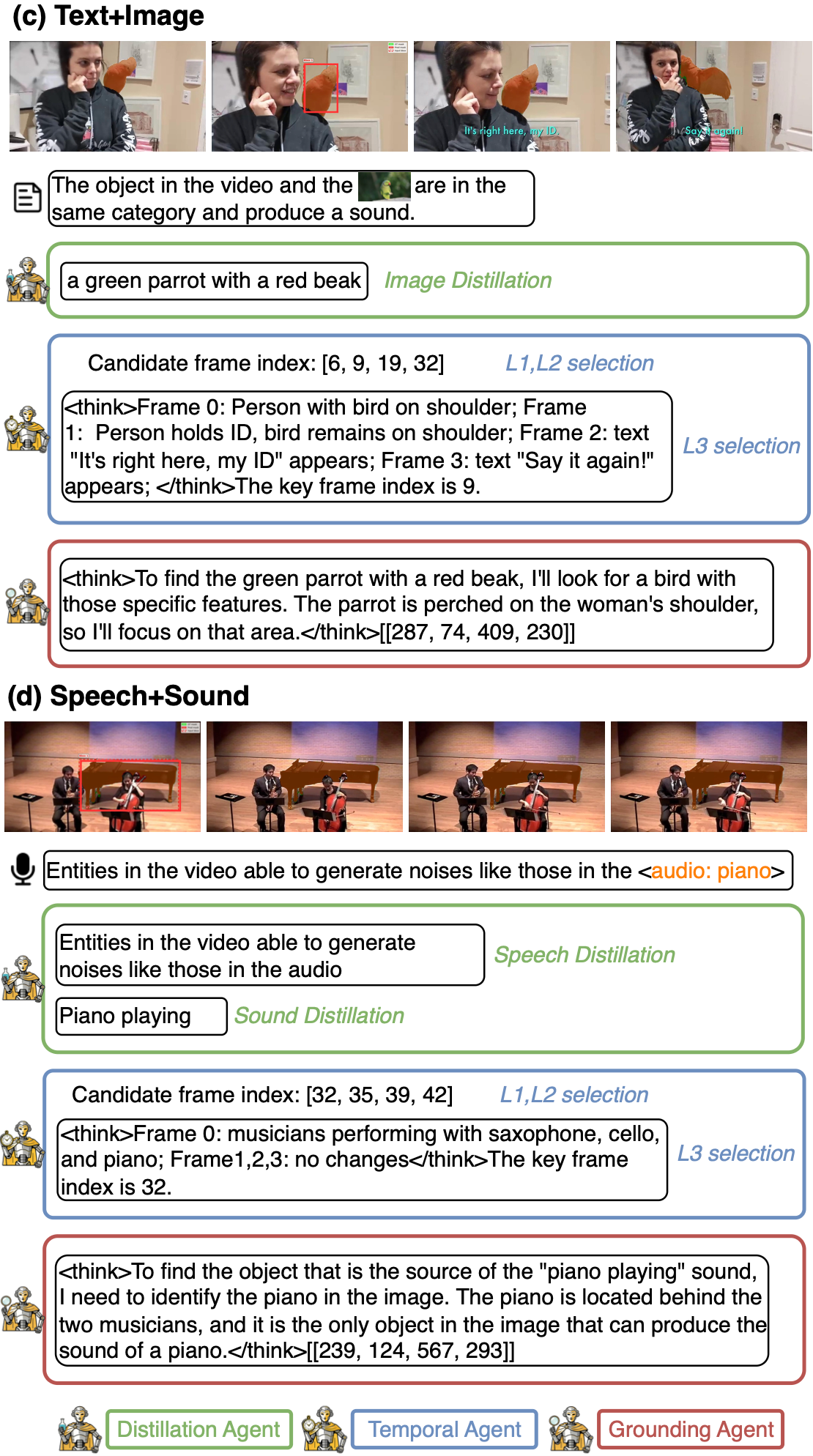}
  \captionof{figure}{Qualitative examples of OPERA on text+image (c) and speech+sound (d) queries.
  In (c), the DA applies image distillation to generate a compact visual description from the reference image.
  In (d), the DA simultaneously produces speech and sound distillation hints, which are merged into a unified grounding query before key frame selection.}
  \label[figure]{fig:suppl-qual-b}
\end{minipage}\par

\par\noindent
\begin{minipage}[t][\fullcolfig][t]{\linewidth}
  \centering
  \includegraphics[width=\linewidth]{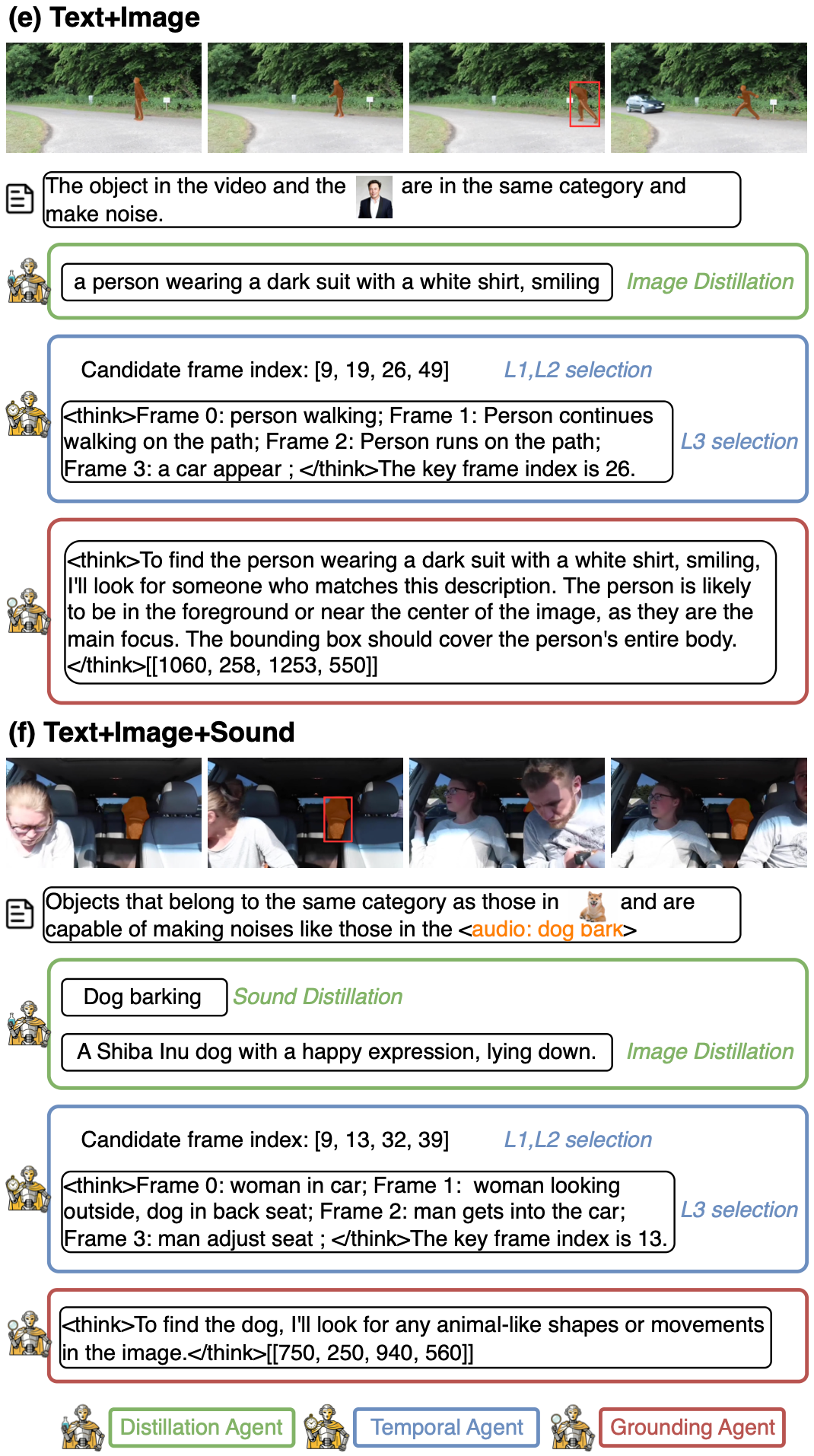}
  \captionof{figure}{Qualitative examples of OPERA on text+image (e) and text+image+sound (f) queries.
  In (e), the DA uses image distillation to identify a person from a reference image, demonstrating cross-modal semantic bridging on a cross-instance retrieval case.
  In (f), both sound and image distillation hints are produced by the DA and jointly inform the grounding query.}
  \label[figure]{fig:suppl-qual-c}
\end{minipage}\par

\end{document}